\documentclass[12pt]{article}

\usepackage[utf8]{inputenc}
\usepackage{newtxtext,newtxmath}
\usepackage{textcomp}

\usepackage{graphicx}
\usepackage{booktabs}
\usepackage[letterpaper,margin=1in]{geometry}
\usepackage{upgreek}
\usepackage{bm}
\usepackage{xurl}
\usepackage{hyperref}
\definecolor{shadowpurple}{RGB}{125,38,205}
\usepackage{wrapfig}
\usepackage{caption}
\usepackage[CaptionAfterwards]{fltpage}
\usepackage[normalem]{ulem}
\renewenvironment{abstract}
	{\quotation}
	{\endquotation}

\date{}

\makeatletter
\renewcommand{\fnum@figure}{\textbf{Figure \thefigure}}
\renewcommand{\fnum@table}{\textbf{Table \thetable}}
\makeatother

\usepackage{scicite}

\usepackage{url}

\def\scititle{
	Robot Learning to Communicate through \\ Projected Visual Abstractions
}
\title{\bfseries \boldmath \scititle}

\author{
	Danyang Yan$^{1\dagger}$,
	Boyuan Wang$^{2\dagger}$,
    Jiaxun Liu$^{2\dagger}$,
    Boyuan Chen$^{1,2,3\ast}$\\
    \normalsize{$^{1}$Department of Electrical and Computer Engineering, Duke University}\\
    \normalsize{$^{2}$Department of Mechanical Engineering and Materials Science, Duke University}\\
    \normalsize{$^{3}$Department of Computer Science, Duke University}\\
    \normalsize{$^\ast$To whom correspondence should be addressed; E-mail: boyuan.chen@duke.edu.}\\
    \normalsize{$^\dagger$These authors contributed equally to this work.}
}

\begin{document} 

\maketitle

\begin{center}
{
\hypersetup{pdfborder={0 0 0}}
\href{https://generalroboticslab.com/shadow}
{\textcolor{shadowpurple}{\texttt{Project website: generalroboticslab.com/shadow}}}
}
\end{center}
\begin{abstract} \bfseries \boldmath
Humans routinely communicate through abstractions of their bodies, including shadows, silhouettes, and reflections. Yet robots remain largely confined to expressing themselves through their physical morphology. Enabling robots to communicate through such projected visual abstractions requires reasoning not only about bodily motion but also about how that motion is transformed into an external representation perceived by an observer. Among these abstractions, shadows provide a particularly compelling example because they emerge directly from the robot's embodiment while remaining visually distinct from the body itself. Here, we present a robotic system capable of dynamic shadow expression using a 21-degree-of-freedom dexterous hand with compliant soft skin and a learned shadow self-model. The soft-skinned embodiment reduces light leakage to produce visually continuous silhouettes, while the differentiable self-model learns the mapping between hand configurations and projected shadow appearance through task-agnostic self-exploration. Given a target shadow image or video, the robot optimizes its hand configurations through gradient-based search over the learned self-model and refines the solution through collision-aware simulation to obtain physically feasible motions. For dynamic shadow performance, we further introduce expressive-region objectives, temporal smoothness regularization, and keyframe-based optimization to preserve visually important motion cues while reducing optimization complexity. We demonstrate robotic shadow expression across sign-language gestures, hand-shadow puppetry, and animal motion imitation in both simulation and physical experiments. These results establish a framework for enabling robots to manipulate projected visual abstractions of themselves for communication and visual storytelling.
\end{abstract}

\section*{Introduction}

Humans routinely communicate through abstraction~\cite{deacon1998symbolic,donald1993origins,tomasello2010origins}. Beyond spoken language and direct physical actions, people convey meaning through symbolic representations such as drawings~\cite{fan2023drawing,fan2020pragmatic,tversky2019mind}, gestures~\cite{mcneill1992hand,goldin2013gesture}, silhouettes, shadows~\cite{stoichita1997short,casati2007shadows}, and visual metaphors~\cite{lakoff2008metaphors}. These representations allow observers to infer information about an object, person, action, and even emotions without directly observing the physical entity itself. We refer to this class of representations as \textit{projected visual abstractions}: visual forms that emerge from, but are distinct from, the underlying physical embodiment. Among these abstractions, shadow performance is particularly intriguing because shadows arise directly from the geometry and motion of a body while transforming it into a fundamentally different visual representation. A performer can intentionally manipulate a three-dimensional morphology, such as a hand, puppet, or body, to generate a desired two-dimensional silhouette that can be recognized and interpreted by an observer.

Humans frequently use shadows, reflections, mirrors, and silhouettes to understand and communicate aspects of their physical presence. Likewise, many animals have been shown to react to shadows and projected body representations in their environments, using them as cues for navigation, communication, and threat detection~\cite{schiff1962looming,yilmaz2013looming,oliva2007crab,schleidt2011hawk}. Unlike direct observation of a body, projected representations are inherently transformed by geometry, lighting, and viewpoint. Consequently, generating a desired shadow is not simply a motor control problem. Instead, it requires reasoning about how an observer perceives the consequences of one's embodiment through projection. In this sense, shadow expression is conceptually related to self-modeling~\cite{bongard2006resilient,kwiatkowski2019task,chen2022fully, hu2025egocentric}, inverse graphics~\cite{kulkarni2015deep}, visual imagination~\cite{ha2018recurrent,Hafner2020Dream}, and perspective-taking~\cite{trafton2005enabling,fischer2016markerless, chen2021visual, chen2021visualtob, ji2025enabling}, where an agent must predict how internal bodily states manifest as external observations.

Developing robots capable of intentionally controlling such abstractions may therefore provide future avenues for communication, storytelling, entertainment, and human-robot interaction. More broadly, it raises a fundamental question: can a robot learn to reason about and manipulate projected representations of itself to convey information, intent, or expression?

Robotic systems have increasingly demonstrated the ability to express themselves through artistic and socially meaningful behaviors. Previous work has enabled robots to paint~\cite{aguilar2008robotic}, perform calligraphy~\cite{sun2013calligraphy}, dance, exhibit stylized locomotion~\cite{song2015development,hopkins2024interactive,muller2025olaf}, and generate expressive facial behaviors~\cite{chen2021smile,hu2024human,hu2026learning}. These systems allow robots to communicate intent, personality, emotion, or aesthetic content through direct observation of their bodies. However, existing approaches primarily focus on expression through the robot’s physical morphology itself. In contrast, shadow expression represents a fundamentally different form of expression in which communication occurs through a projected representation of the body. To our knowledge, robots have not previously been enabled to intentionally generate and control shadows as a medium of dynamic expression.

Shadow-based communication has a rich history in human culture. Since prehistoric cave art, silhouettes have been used to depict animals, people, and narratives~\cite{fernandez2026reevaluating,leroi1982dawn}. Over time, shadows evolved into sophisticated artistic and performance media~\cite{currell2015shadow}. Shadow puppetry remains one of the oldest forms of visual storytelling~\cite{chen2003shadow,chen2007chinese,zhang2012chinese}, while hand-shadow art demonstrates how complex and recognizable forms can emerge from dexterous manipulation of the human hand~\cite{almoznino2002art}. More recently, shadows have been explored in painting~\cite{forgione1999shadow}, digital media~\cite{tsuchiya2019virtual}, computational fabrication~\cite{mitra2009shadow,bermano2012shadowpix}, and optimized static shadow sculptures~\cite{won2016shadow,xu2025hand}. These works demonstrate the expressive power of shadows, yet they largely rely on static geometries or human performers.

Unlike static sculptures, robots can continuously adapt their morphology to generate a diverse range of shadows and dynamic shadow narratives. However, enabling a robot to perform shadow art introduces several challenges that differ substantially from conventional motion generation and visual imitation tasks. A central challenge is that shadow expression requires a robot to reason about the relationship between embodiment and appearance. Similar to prior work on robot self-modeling~\cite{chen2022fully}, robots must learn how changes in their internal configurations alter external visual representations. Self-models have previously enabled robots to reason about their own morphology, predict future body states, recover from damage, and generate facial expressions~\cite{chen2022fully,hu2024human,hu2026learning}. However, unlike traditional self-models that predict body geometry, occupancy, or facial deformations, shadow expression requires reasoning over a many-to-one projection from three-dimensional morphology to a two-dimensional silhouette. Multiple physically distinct hand configurations may produce nearly identical shadows, making the inverse problem fundamentally ambiguous. This ambiguity becomes particularly challenging during motion generation, where temporally coherent shadow transitions must be achieved despite the existence of many possible solutions for each frame.

\begin{figure}[t!]
    \centering
    \includegraphics[width=0.8\textwidth]{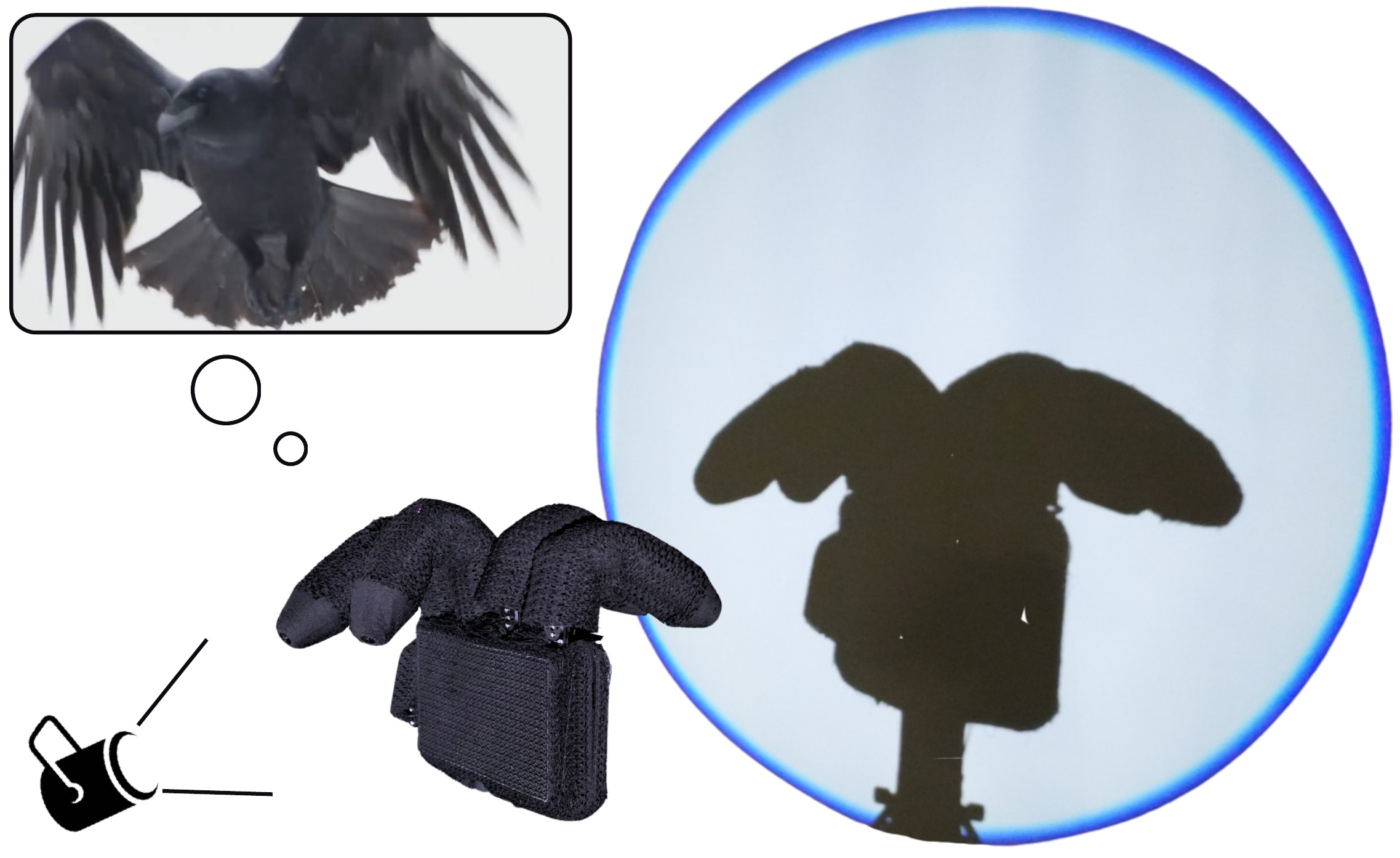}
    \caption{$|$ \textbf{A robot hand actively generating expressive shadow performances through learned shadow self-modeling.}}
    \label{fig:teaser}
\end{figure}

A second challenge arises from the physical embodiment itself. Human hand-shadow artists rely heavily on soft tissue interactions and continuous surface contact between fingers to block light and create visually coherent silhouettes. In contrast, most robotic hands are constructed from rigid linkages separated by gaps that accidentally allow light leakage. These gaps, though seemingly subtle, can substantially degrade shadow quality and limit the range of achievable silhouettes. Therefore, enabling expressive robotic shadow performance requires not only a computational framework for shadow reasoning but also hardware innovation capable of forming visually continuous projections.

A third challenge concerns physical feasibility. Even if a shadow can be reproduced through optimization, the resulting joint configuration may contain self-collisions or mechanically infeasible poses. Furthermore, when reproducing dynamic targets such as animal motions or shadow puppetry performances, the robot must preserve temporally coherent motion while simultaneously reproducing visually important details that contribute disproportionately to the perceived expressiveness of the shadow.

In this work, we enable robots to perform expressive shadow art through a dexterous robotic hand (fig.~\ref{fig:teaser}). We developed a 21-degree-of-freedom (DoF) robotic hand composed of a rigid skeletal structure and compliant soft skin. The soft outer layer provides human-hand-like sealing capability that substantially reduces light leakage during shadow formation while preserving dexterous finger motion and allowing for deformable finger contacts. Unlike conventional rigid robotic hands with visible gaps between linkages, our compliant design enables the generation of visually continuous and expressive shadow silhouettes.

To control this robotic hand for shadow generation, we first train a shadow self-model that establishes a fully differentiable mapping from joint configurations to the resulting 2D shadow appearance. Rather than directly predicting joint angles from target shadows, which is inherently ambiguous due to the many-to-one relationship between hand poses and silhouettes, our framework learns a forward shadow self-model through task-agnostic self-exploration in simulation. Given a single static target shadow image, the robot backpropagates the reconstruction error through the learned shadow self-model to optimize the hand pose and imitate the target shadow.

However, optimization through the shadow self-model alone often produces physically infeasible configurations. To address this challenge, we use the optimized pose as a warm start for a collision-aware search process within a physics simulator. This second-stage optimization refines the solution into a physically valid and executable hand configuration. This hybrid strategy combines the efficiency of differentiable optimization with the physical realism of simulation-based refinement, enabling reliable sim-to-real shadow reproduction.

When imitating moving targets with more complex silhouette shapes and transitions, such as videos of moving animals or hand-shadow puppetry, the task becomes increasingly challenging due to the need for both temporal smoothness across video frames and detailed shape reproduction. Many expressive visual cues arise not only from the overall silhouette, but also from subtle moving or enclosed regions, such as the motions of wings, the eyes of an animal, or the shape of a beak. By identifying and emphasizing these regions during optimization, our robot can preserve details that contribute disproportionately to perceived expressiveness.

Furthermore, we observe that a sparse set of representative keyframes is often sufficient to capture the essential structure and dynamics of a target motion sequence. We therefore develop an iterative clustering method for keyframe extraction from raw target videos that substantially reduces optimization complexity while preserving characteristic motion patterns. Combined with temporal smoothness regularization and expressive-region objectives, our strategy enables the generation of temporally coherent shadow performances from complex video targets.

The ability to imitate both static shadow targets and dynamic video sequences enables our robot to perform a wide variety of shadow expressions, including shadow pictogram displays with hand gestures, hand puppetry shows to imitate animals, and expressive animal motions. Experimental results demonstrate that our framework successfully reproduces both global shadow structures and subtle visual details while maintaining smooth and coherent shadow transitions over time. Overall, our work expands robotic expression into the domain of projected visual abstractions and establishes a framework for future robotic systems capable of communicating through shadows, silhouettes, and other embodiment-dependent visual representations.

\section*{Results}

\begin{figure}[t!]
    \centering
    \includegraphics[width=1\textwidth]{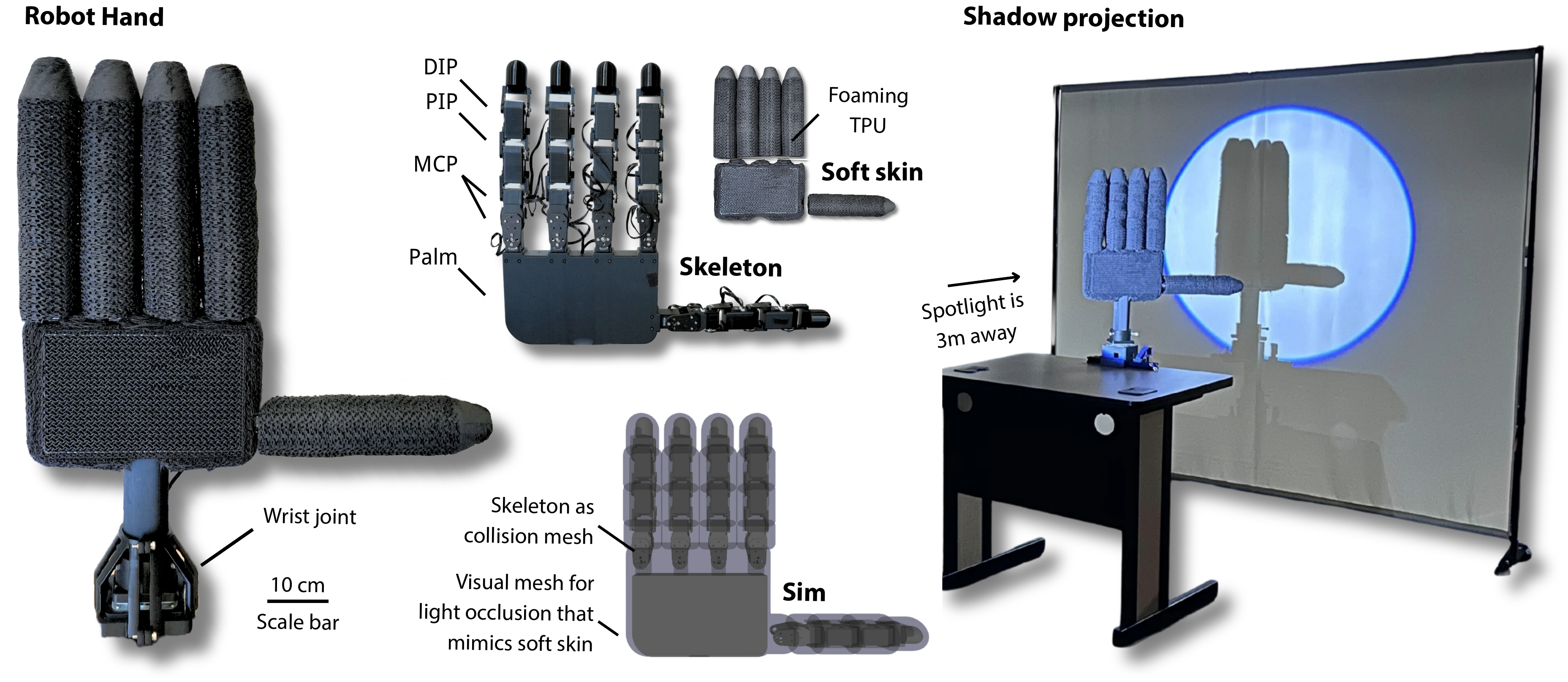}
    \caption{{$|$ \textbf{Robot hand design for expressive shadow communication.} The robot hand consists of a rigid skeletal structure covered by a compliant soft outer skin. The skeleton includes five identical fingers, each composed of distal interphalangeal (DIP), proximal interphalangeal (PIP), and metacarpophalangeal (MCP) joints. Each MCP joint provides two DoF, enabling both flexion–extension and abduction–adduction motions. A wrist joint is located at the base of the hand support. To enable effective light sealing, the skeleton is enclosed by soft skin fabricated using TPU foaming 3D printing. This process produces a lightweight and compliant material that introduces minimal resistance to the motion of the skeleton. In simulation, the skeleton is used as the collision mesh, whereas the visual mesh approximates the soft skin to reproduce light occlusion effects. For real-world shadow projection, the robot hand is mounted on a table with a spotlight positioned 3 meters away. A white backdrop is placed 1 meter behind the hand to capture the projected hand shadow.}}
    \label{fig:hardware}
\end{figure}

\subsection*{Design of a robotic hand for expressive shadow communication}

To communicate through projected visual abstractions, a robot must generate shadows that are both visually recognizable and physically controllable. This requirement introduces a unique embodiment challenge. A conventional rigid robotic hand is poorly suited for shadow expression because gaps between rigid linkages allow light to leak through the fingers. These gaps fragment the projected silhouette and limit the ability of the hand to form coherent shadow shapes. Fully soft robotic hands, although more compliant, are also not ideal for this task because their motion is often less precisely constrained, harder to model, and less directly compatible with high-dimensional pose optimization. Shadow expression requires both reliable dexterous articulation and continuous light-blocking and compliant contact. Human hand-shadow artists achieve this combination through rigid skeletal support covered by soft tissue, allowing fingers to move precisely while sealing light paths through compliant contact. Inspired by this structure, we designed each finger as a slender rigid skeleton covered by a compliant foam layer (fig.~\ref{fig:hardware}). The foam was fabricated by low-infill TPU 3D printing, which allowed the geometry and stiffness of the skin to be customized while introducing minimal resistance to the underlying skeletal motion (see ``Mechanical Design'' section in Materials and Methods). This hybrid rigid-soft design preserved kinematic controllability while enabling neighboring fingers to conform to one another and improve light blocking during shadow formation. Overall, each finger consisted of a four-link kinematic chain approximating the distal, proximal, and metacarpophalangeal joints of a human finger. Each metacarpophalangeal (MCP) joint provided two degrees of freedom (DoFs), enabling both flexion–extension and abduction–adduction motions. Together with an additional wrist joint at the base, the hand provided 21 DoFs.

We evaluated the hand within a projection-based communication setup shown in Figure~\ref{fig:hardware}. The robotic hand was mounted on a table, and a high-intensity spotlight was positioned approximately three meters from the palm center to project the hand shadow onto a white backdrop. The backdrop was placed one meter behind the hand and served as the projection plane. A camera positioned behind the backdrop captured the resulting real shadow images. In simulation, we replicated this setup by placing a white planar mesh behind the hand to visualize shadows (fig.~\ref{fig:framework}A). Collision detection was computed using the slender skeletons, whereas the visual meshes of the links were enlarged to approximate the light-blocking geometry of the soft foam layer. This separation between collision geometry and visual occlusion geometry allowed the simulator to capture both physical feasibility and shadow appearance.

\begin{FPfigure}
    \centering
    \includegraphics[width=1\textwidth]{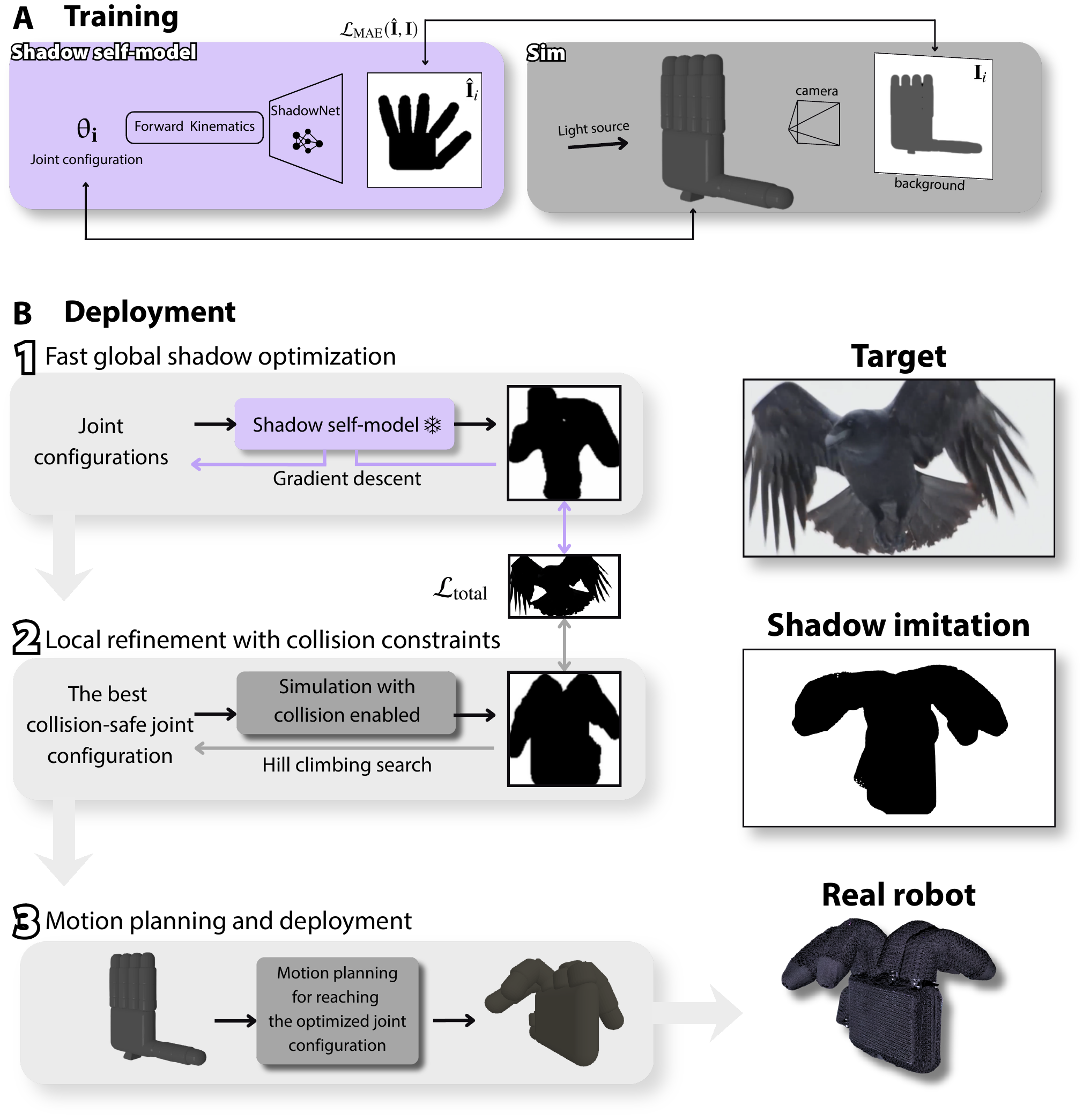} 
	\caption{{$|$ \textbf{Shadow self-modeling framework.} (\textbf{A}) A shadow self-model was learned in simulation, where the real-world hand projection setup was replicated virtually. The robot hand was commanded with random joint configurations (i.e., ``motor bubbling''), and the corresponding shadow images were recorded to form a training dataset. The shadow self-model consisted of an analytical forward kinematics model followed by a neural network. Overall, the model took joint configurations $\boldsymbol{\uptheta}_{i}$ as input and predicted the corresponding binary shadow images $\hat{\mathbf{I}}_{i}$. The mean absolute error(MAE) loss was computed against the ground-truth images ${\mathbf{I}}_{i}$, and backpropagated to optimize the neural network parameters.
    (\textbf{B}) During deployment, a target shadow image was first binarized. In the first phase, the trained shadow self-model was frozen, and random joint configurations were sampled as inputs to the model. The loss between the predicted shadow and the target binary shadow image was then backpropagated to optimize the joint angles. The best generated joint configuration was further refined through a hill-climbing search in simulation that used the same projection setting as in training. With collision checking enabled, this process produced a collision-free and physically feasible pose for the real robotic hand. The optimized joint configuration was then used as the final target for the physical hand, and motion planning was performed in simulation to generate a collision-free trajectory. The real robotic hand executed this motion plan to produce the final shadow imitation result.}}
    \label{fig:framework}    
\end{FPfigure}

\subsection*{Targets for projected visual abstraction}

We evaluated shadow communication across static and dynamic target shadows with increasing levels of difficulty. The dataset contains 26 static hand gesture images from American Sign Language (ASL) and image frames extracted from six video targets. The video targets include four hand-performed animal shadow puppetry videos, featuring a duck, deer, peacock, and camel, and two raw animal videos depicting a howling wolf and a raven flapping its wings. The full hand gesture images and videos are provided in Supplementary Material Figures~\ref{fig:full_alphabet}–~\ref{fig:full_wolf} and Supplementary Movies 1-7. From these video targets, we extracted 19 representative keyframes from the shadow puppetry videos and 16 representative keyframes from the raw animal videos (fig.\ref{fig:clustering_and_keyframe} and~\ref{fig:rest_of_key_frames}; see “Keyframe Extraction for Video
Targets” in Materials and Methods). In total, the evaluation contains 61 image targets.

These targets were selected to evaluate the framework across progressively larger embodiment gaps between the target silhouette and the robotic hand. The 26 ASL hand gesture targets represent the most direct setting, where the target silhouettes are kinematically feasible and closely aligned with the morphology of our robot hand. The 19 animal shadow puppetry targets are more challenging because they involve dynamic hand-generated animal silhouettes that are thinner, more stylized, and structurally more complex than the ASL targets. These targets test whether our framework can reproduce expressive shadow motion despite differences between human and robotic hand morphology. Finally, the 16 raw animal video targets represent the most challenging setting because the target silhouettes do not directly correspond to any hand embodiment. These targets test whether our framework can generate visually meaningful approximations of non-hand shapes and motions through its own projected morphology. We note that our framework does not require any training data from human or existing video datasets. All training solely relies on the robot's own data in a self-supervised manner.

Our evaluations include both quantitative metrics and qualitative comparisons between target shadows and generated shadows. The optimization objective combines pixel-level reconstruction, silhouette overlap, and perceptual similarity through mean absolute error (MAE), intersection-over-union (IoU) loss, and cosine similarity between Contrastive Language-Image Pre-training (CLIP) embeddings~\cite{zhang2022contrastive,radford2021learning}. Detailed objective functions and optimization settings are provided in the following sections.

\subsection*{Learning a shadow self-model}

To enable shadow expression, our robot first learned how its joint configurations transform into projected shadow appearances. This mapping is difficult to invert directly because many physically distinct hand poses can produce identical or highly similar two-dimensional silhouettes. As a result, directly training an inverse model from target shadow images to joint configurations is ill-posed. As shown in Table~\ref{tab:single_image_results}, the inverse baseline performs poorly and remains close to random guessing from the training dataset.

We therefore trained a forward shadow self-model that predicts shadow appearance from joint configuration. The forward model provides a differentiable mapping from a 21-dimensional joint angle vector to a two-dimensional binary shadow image (Fig.~\ref{fig:framework}A). The challenge in this forward model is that it must simultaneously learn the relationship between joint configurations and the resulting 3D hand morphology, as well as the projection of this 3D morphology into a 2D shadow image. Our model hence incorporates an analytical forward kinematics module that transforms joint angles into structured three-dimensional transformations of the hand joints, followed by a neural network that predicts the resulting projected silhouette. We found that our design gave the network explicit kinematic structure before learning the remaining projection and occlusion effects. Compared with a neural network that directly maps joint angles to shadow images, including analytical forward kinematics improved performance by 31.08\% when evaluated on the 26 hand gesture targets. Without this structured kinematic representation, the network must implicitly learn both hand geometry and projection, which produces a less reliable gradient landscape for downstream optimization.

\begin{figure}[t!]
    \centering
    \includegraphics[width=\textwidth]{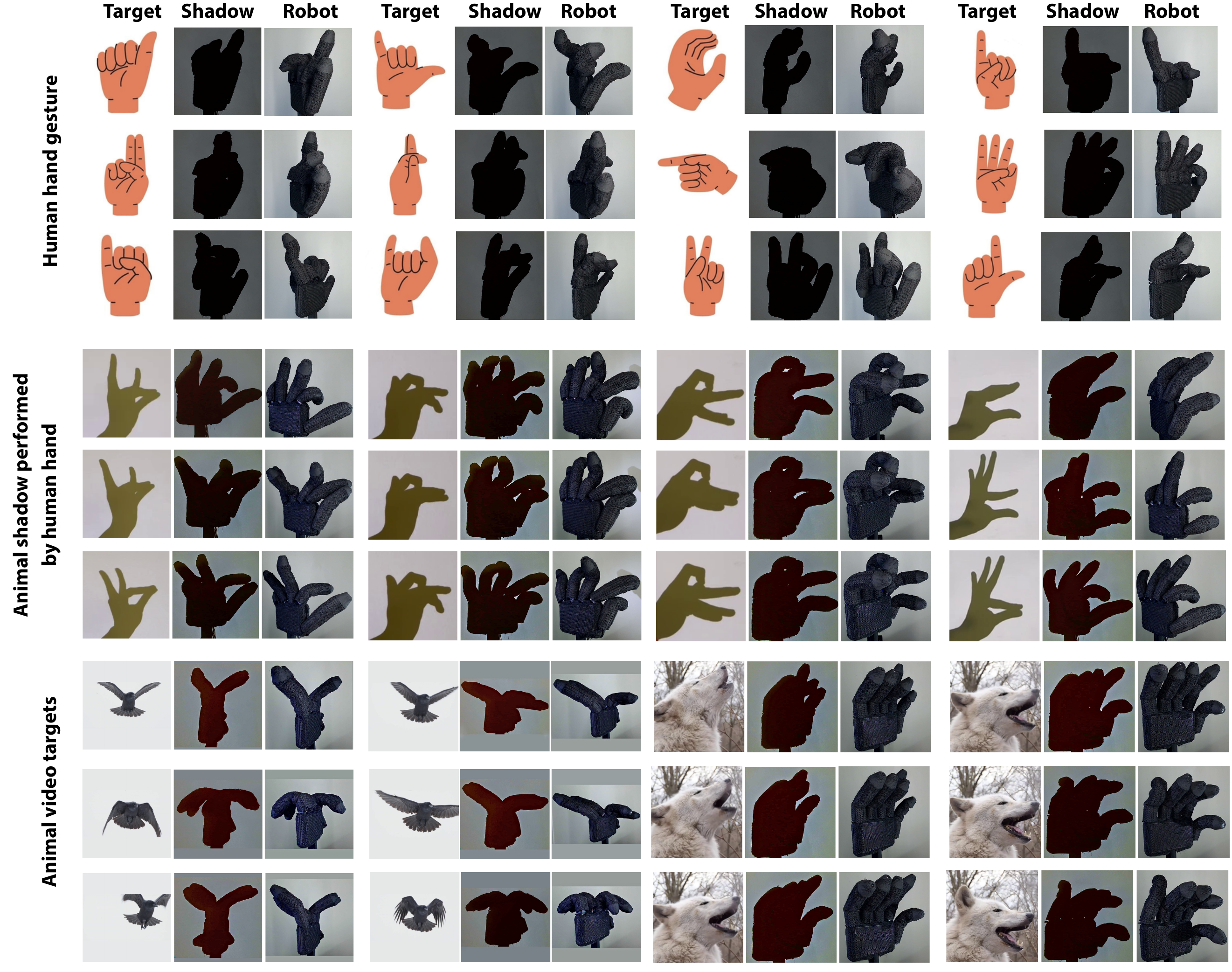}
   \caption{{$|$ \textbf{Shadow imitation results for static image targets and video frame targets.} Selected examples (36 of 61 targets) demonstrate shadow optimization from static hand gesture images, hand-performed animal shadow puppetry frames, and raw animal video frames. Full results are provided in the Supplementary Materials. Our robot captures visually distinct styles across targets and reproduces both global silhouette structures and fine details, such as eyes, slender fingers of the mouse, and animal ears.}}
    \label{fig:fig_shadow_human_hand_target}
\end{figure}

Our robot learned this shadow self-model entirely in simulation through task-agnostic self-exploration. During data collection, the hand randomly sampled joint configurations and observed the corresponding projected shadows (see ``Shadow imitation framework – Data collection’’ in Materials and Methods). The resulting model achieved a test MAE of 0.00643 on unseen joint configurations using the same sampling strategy. Qualitative predictions are shown in Figure~\ref{fig：shadow self-model qualitative}.

During deployment, the shadow self-model was frozen and used as a differentiable surrogate for shadow generation. We initialized a batch of joint configurations $\boldsymbol{\uptheta}\in \mathbb{R}^{21}$ and predicted the corresponding shadow images using $\hat{\mathbf{I}} = S_{\phi}(\boldsymbol{\uptheta})$. Given a target shadow image $\mathbf{I}^{\text{target}} \in \mathbb{R}^{256 \times 256}$, we optimized the joint configuration by minimizing a weighted combination of MAE, IoU loss, and CLIP embedding distance:
\begin{equation}
\mathcal{L}_{\text{base}} =
\lambda_{\text{MAE}} \, \| \mathbf{I}^{\text{target}} - \hat{\mathbf{I}} \|_1
+ \lambda_{\text{IoU}} \, \left( 1 - \mathrm{IoU}(\mathbf{I}^{\text{target}}, \hat{\mathbf{I}}) \right)
+ \lambda_{\text{CLIP}} \, \left(
1 - 
\frac{\mathbf{f}^{\text{pred}}}{\|\mathbf{f}^{\text{pred}}\|_2}
\cdot
\frac{\mathbf{f}^{\text{target}}}{\|\mathbf{f}^{\text{target}}\|_2}
\right),
\end{equation}
where $\mathbf{f}^{\text{pred}}$ and $\mathbf{f}^{\text{target}}$ denote the CLIP image embeddings of the predicted and target images, respectively. The MAE term encourages pixel-level accuracy, the IoU term emphasizes overlap between foreground silhouettes, and the CLIP term provides a more flexible perceptual matching signal. We then optimized
\begin{equation}
\boldsymbol{\uptheta}^* = \arg\min_{\boldsymbol{\uptheta}} \mathcal{L}_{\text{total}}(\boldsymbol{\uptheta}).
\end{equation}
through gradient descent over the input joint configuration while keeping the self-model fixed (see ``Shadow imitation framework – Optimization’’ in Materials and Methods).

\begin{table}[t]
\centering
\caption{Quantitative results for all 61 single-image targets. $\downarrow$ indicates lower is better, while $\uparrow$ indicates higher is better. The same notation is used throughout the paper. The random baseline randomly selects a shadow image from the training dataset as the prediction. The inverse baseline uses a model that takes a target shadow image as input and directly predicts the corresponding joint angles. The predicted joint configuration is then rendered in simulation, and the loss is computed between the target shadow and the resulting simulated shadow. The nearest-neighbor baseline selects the shadow image in the training dataset with the lowest base loss relative to the target image to assess generalization beyond memorization of the training data. H.C. denotes hill-climbing optimization. Total loss and base loss are weighted, whereas the remaining losses are reported as raw values.}
\label{tab:single_image_results}

\setlength{\tabcolsep}{4pt}

\resizebox{\columnwidth}{!}{
\begin{tabular}{lccccccc}
\noalign{\hrule height 1.5pt}
Method & Total$\downarrow$ & Base$\downarrow$ & CLIP$\downarrow$ & MAE$\downarrow$ & IoU Loss$\downarrow$ & Exp. IoU$\downarrow$ & Exp. CLIP$\downarrow$ \\
\hline
Random & 0.4590 & 0.2135 & 0.1116 & 0.3402 & 0.4772 & 0.2434 & 0.0414 \\
Inverse & 0.4262 & 0.1959 & 0.1246 & 0.3102 & 0.4370 & 0.2286 & 0.0337 \\
Nearest Neighbor & 0.2014 & 0.0849 & 0.0782 & 0.1160 & 0.1910 & 0.1155 & 0.0192 \\
Ours (Base without H.C.) & 0.1809 & 0.0828 & \textbf{0.0681} & 0.1090 & 0.1873 & 0.1422 & 0.0233 \\
Ours (Base) & \textbf{0.1210} & \textbf{0.0640} & 0.0707 & \textbf{0.0829} & \textbf{0.1440} & \textbf{0.0863} & \textbf{0.0182} \\
\noalign{\hrule height 1.5pt}
\end{tabular}
}

\end{table}

\subsection*{From projected appearance to embodied execution}

Optimizing solely through the shadow self-model produced joint configurations that match target silhouettes (Ours (without H.C.) in fig. ~\ref{fig:qualitative}A and C), but these configurations were not always physically executable. The learned self-model captured projected appearance but did not explicitly enforce physical constraints such as inter-finger collisions. Consequently, the optimized joint configuration could conflict with the constraints of the robot embodiment. Across the 61 targets, 39.34\% of the shadow self-model-optimized poses exhibited collision issues and could not be directly deployed on the real robotic hand.

To reconcile projected appearance with physical embodiment, we introduced a second optimization stage in a physics-enabled simulator (Fig.~\ref{fig:framework}B). After gradient-based optimization through the shadow self-model, the best joint configuration was used as a warm start for a local hill-climbing search in simulation. This simulator used the same projection setup as training while incorporating collision checking. The search was restricted to small random perturbations around the self-model solution, allowing the method to refine the pose for physical feasibility while preserving the target shadow appearance (see ``Shadow imitation framework – Optimization’’ in Materials and Methods).

This hybrid strategy improved both performance and deployability. As shown in Table~\ref{tab:single_image_results}, adding hill-climbing refinement (Ours(Base)) improved the total loss by 33.1\% relative to the self-model-only result (Ours(Base without H.C.)). Pure hill climbing is computationally expensive and prone to poor local minima in the high-dimensional joint space because it relies only on random local exploration. In contrast, our method first used differentiable optimization to efficiently identify a high-quality region of the configuration space, and then used physics-based local search only to refine the solution for feasibility. As shown in Figure~\ref{fig:qualitative}B, 500 steps of hill climbing initialized from the self-model solution outperformed 500 steps of pure hill climbing. Although 2000 steps of pure hill climbing slightly improved performance, it substantially increased runtime and still performed worse than the hybrid method. Qualitatively, pure hill climbing could reproduce coarse shadow shapes but often failed to capture fine structures such as eyes and thin regions, and could introduce small artifacts. In contrast, the hybrid approach preserved both global silhouette structure and subtle but important visual details (fig.~\ref{fig:qualitative}A and C).

After a physically feasible target configuration was obtained, we employed a sample-based trajectory planner to generate a collision-free motion plan from the current hand configuration to the optimized target pose (see ``Motion planning for sim-to-real transfer’’ in Materials and Methods). For sequential imitation, the planner first moved the robot from the home configuration to the first target pose and then planned between consecutive target configurations to avoid collisions. The planned trajectories were executed directly on the physical robot hand to produce real-world shadow targets (see Supplementary Movies 1-7).

\subsection*{Shadow imitation across embodiment gaps}

We first evaluated the complete imitation framework in simulation across all 61 targets by treating each target independently as a single-image imitation task. As shown in Table~\ref{tab:single_image_results}, our method outperformed all baseline approaches, including random selection, inverse prediction, nearest-neighbor retrieval, and self-model optimization without physics-based refinement. The results showed that forward shadow self-modeling, followed by collision-aware refinement, provided an effective strategy for mapping target silhouettes to executable hand configurations. For targets that closely matched the robot embodiment, such as ASL hand gestures, our framework successfully reproduced target shadow shapes on the physical robot hand (fig.~\ref{fig:fig_shadow_human_hand_target} and fig.~\ref{fig:full_alphabet}). These targets were kinematically compatible with the robot hand and therefore primarily evaluated whether the system could accurately solve the inverse shadow problem.

The task became substantially more challenging when the target silhouettes differed from the robot embodiment (fig.~\ref{fig:fig_shadow_human_hand_target}, fig.~\ref{fig:expressive_region_demonstration}). Hand-performed animal shadow puppetry often uses thinner human fingers and more delicate inter-finger openings than the robot can physically reproduce, whereas raw animal videos contain structures such as raven wings, wolf jaws, and animal ears that do not correspond directly to a hand morphology. Because the targets were represented only as binary silhouettes, direct re-targeting was ambiguous and could not rely on semantic correspondences between target body parts and robot fingers. Despite these embodiment mismatches, our method remained capable of capturing both global shadow structures and important local details across hand gesture, shadow puppetry, and raw animal targets, as shown in the physical demonstrations in Figure.~\ref{fig:fig_shadow_human_hand_target}.

\subsection*{Shadow imitation for dynamic video targets}

\begin{figure}[t!]
	\centering
	\includegraphics[width=1\textwidth]{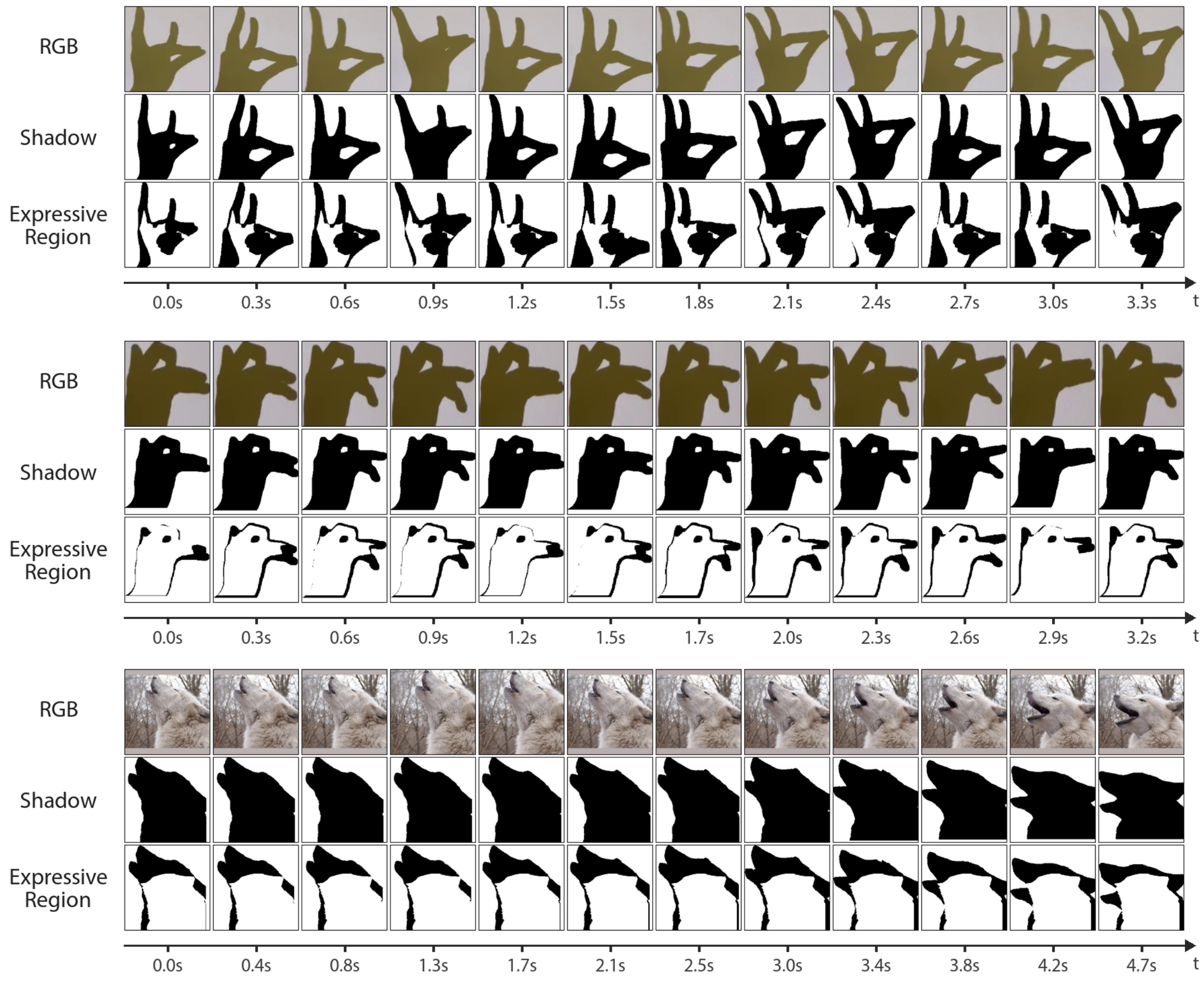}
	\caption{\textbf{Examples of extracted video frames and their expressive regions.} In video target imitation, expressive regions guide the optimization toward areas that contain informative and visually expressive features, such as enclosed white regions in the deer example and moving regions across video sequences.}
	\label{fig:expressive_region_demonstration}
\end{figure}

While single-image shadow imitation evaluates whether our robot can reproduce a desired projected appearance, dynamic shadow targets introduce additional challenges. Beyond matching individual silhouettes, the robot must preserve visually important details, maintain temporally coherent motion, and efficiently reproduce long sequences of shadow transformations. We therefore investigated whether our robot could communicate through dynamic projected visual abstractions derived from hand-shadow puppetry and animal motion videos.

A first challenge arises from the observation that not all regions of a shadow contribute equally to perception. In dynamic shadow performances, expressive information is often concentrated in small localized structures, such as eyes, mouths, beaks, wing boundaries, or other enclosed regions within the silhouette. In addition, motion itself provides information about which portions of a shadow are visually important. Optimizing only for global silhouette similarity can therefore reproduce coarse shape while failing to preserve details that strongly influence perception.

To address this issue, we introduced an expressive-region objective that emphasized visually informative regions during optimization (see ``Expressive Region’’ in Materials and Methods). Expressive regions were identified from two sources: moving shadow regions across consecutive video frames and enclosed empty regions within the silhouette, such as eyes (fig.~\ref{fig:expressive_region_demonstration}). For each frame $i$ of a video target, these regions were used to construct spatial masks $\mathbf{M}_i \in \{0,1\}^{H \times W}$ that highlighted areas expected to carry expressive information. Additional region-weighted IoU and CLIP losses were then incorporated into the optimization objective to emphasize reconstruction accuracy within these expressive regions:
\begin{equation}
\mathcal{L}_{\text{expr}} =
\lambda_{\text{Exp\_IoU}}
\left(
1 -
\mathrm{IoU}
\left(
\mathbf{M} \odot \mathbf{I}^{\text{target}},
\mathbf{M} \odot \hat{\mathbf{I}}
\right)
\right)
+
\lambda_{\text{Exp\_CLIP}}
\left(
1 -
\frac{\mathbf{g}^{\text{pred}}}{\|\mathbf{g}^{\text{pred}}\|_2}
\cdot
\frac{\mathbf{g}^{\text{target}}}{\|\mathbf{g}^{\text{target}}\|_2}
\right),
\end{equation}
where $\odot$ denotes element-wise multiplication and $\mathbf{g}^{\text{pred}}$ and $\mathbf{g}^{\text{target}}$ denote CLIP embeddings of the masked predicted and target images, respectively.

\begin{figure}[t!]
    \centering
    \includegraphics[width=1\textwidth]{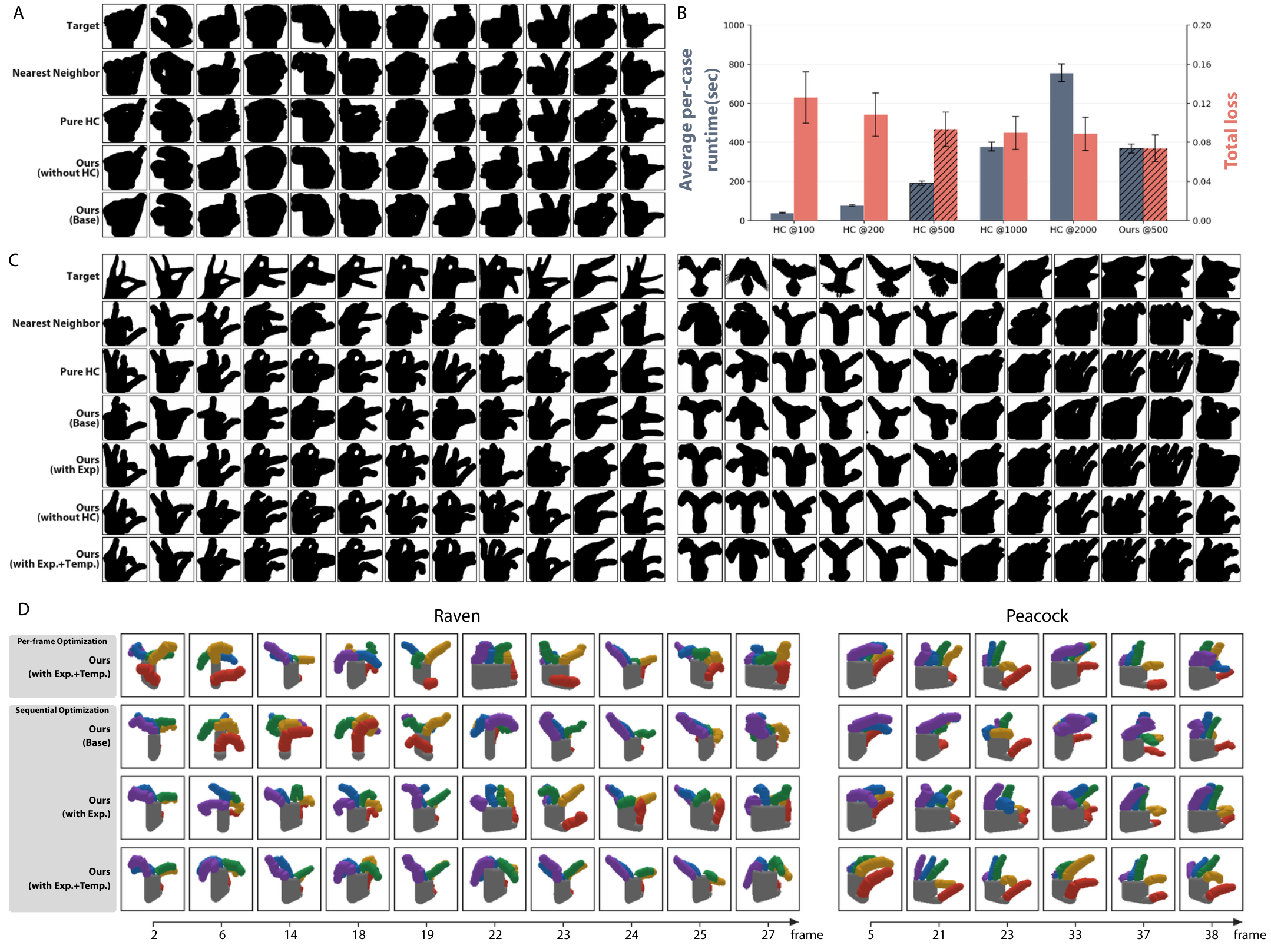}
    \caption{{$|$ \textbf{Qualitative results of ablation studies and hill-climbing analysis.} 
    (\textbf{A}) Twelve ASL examples demonstrating qualitative results of the ablation methods. 
    (\textbf{B}) Comparison between our base method and pure hill-climbing optimization in computation cost and imitation performance. 
    (\textbf{C}) Twelve qualitative results from keyframes extracted from human hand animal shadow puppetry videos. 
    (\textbf{D}) Examples comparing the smoothness of finger transitions between frames. Colors track the motion of each finger: red, yellow, green, blue, and purple correspond to the thumb, index, middle, ring, and pinky fingers, respectively.}}
    \label{fig:qualitative}
\end{figure}

We found that emphasizing expressive regions improved both the motion fidelity and perceptual quality through the preservation of visually salient structures and motion characteristics. As shown in Table~\ref{tab:ablation_results} - Independent Per-frame Optimization, introducing the expressive-region objective substantially reduced expressive-region IoU loss and expressive-region CLIP loss. Although this objective slightly increased the global reconstruction loss, it shifted optimization toward perceptually meaningful details. Qualitatively, the resulting shadows more accurately reproduced subtle but important features such as the eyes of the duck, enclosed regions of the deer, and fine structural details of the animal silhouettes (Fig.~\ref{fig:qualitative}).

\begin{table}[t]
\centering
\caption{Quantitative results of the ablation study evaluating the effects of expressive-region objectives, inherited initialization, and temporal consistency on 35 sequential target images extracted from video sequences. $\downarrow$ indicates lower is better, while $\uparrow$ indicates higher is better. The base loss is computed as a weighted combination of CLIP, MAE, and IoU loss. The total loss is computed as the sum of the base loss and the weighted expressive-region loss. Weight settings are provided in ``Shadow imitation framework – Optimization'' in Materials and Methods.}
\label{tab:ablation_results}

\setlength{\tabcolsep}{4pt}

\resizebox{\columnwidth}{!}{
\begin{tabular}{lcccccccc}
\noalign{\hrule height 1.5pt}
Method & Total$\downarrow$ & Base$\downarrow$ & CLIP$\downarrow$ & MAE$\downarrow$ & IoU Loss$\downarrow$ & Exp. IoU$\downarrow$ & Exp. CLIP$\downarrow$ &
Transition S.R.\\
\hline

\multicolumn{9}{c}{\textbf{Independent Per-frame Optimization}} \\
\hline

Ours (Base) & 0.1996 & \textbf{0.0861} & 0.0948 & \textbf{0.1039} & \textbf{0.1950} & 0.1618 & 0.0330 & 10/29(34.5\%) \\

Ours (with Exp.) & \textbf{0.1781} & 0.1141 & \textbf{0.0871} & 0.1577 & 0.2572 & \textbf{0.0628} & 0.0235 & 5/29(17.2\%) \\

Ours (with Exp.+ Temp.) & 0.1789 & 0.1153 & 0.0891 & 0.1605 & 0.2596 & 0.0625 & \textbf{0.0215} & 5/29(17.2\%)\\

\hline
\multicolumn{9}{c}{\textbf{Sequential Optimization with Inherited Initialization}} \\
\hline

Ours (Base) & 0.2684 & \textbf{0.0969} & 0.0858 & 0.1220 & \textbf{0.2196} & 0.1701 & 0.0287 & 14/29(48.3\%)\\

Ours (with Exp.) & 0.2070 & 0.1219 & 0.0934 & 0.1699 & 0.2747 & \textbf{0.0839} & 0.0231 & 21/29(72.4\%)\\

Ours (with Exp.+ Temp.) & \textbf{0.1871} & 0.1246 & \textbf{0.0685} & \textbf{0.1073} & 0.2920 & 0.1082 & \textbf{0.0227} & 25/29(86.2\%) \\

\noalign{\hrule height 1.5pt}
\end{tabular}
}
\end{table}

A second challenge arises from temporal consistency. Independently optimizing each frame can produce visually similar shadows while assigning different fingers to represent the same semantic shadow region across consecutive frames. As a result, the generated shadows may appear correct when viewed frame-by-frame but produce unstable transitions when executed by the physical robot. For example, during raven-wing imitation, one finger may represent a particular wing boundary in one frame but a different finger may assume that role in the next frame. Such changes require large physical motions between frames and generate unintended intermediate shadows (fig.~\ref{fig:qualitative}D).

To encourage temporally coherent solutions, we introduced a temporal regularization term that penalized deviations from the optimized configuration of the previous frame:
\begin{equation}
\mathcal{L}_{\text{temp}}(\boldsymbol{\uptheta}_i)
=
\lambda_{\text{temp}}
\left\|
\boldsymbol{\uptheta}_i
-
\boldsymbol{\uptheta}_{i-1}^{*}
\right\|_2^2 .
\end{equation}

\begin{figure}[t!]
    \centering
    \includegraphics[width=1\textwidth]{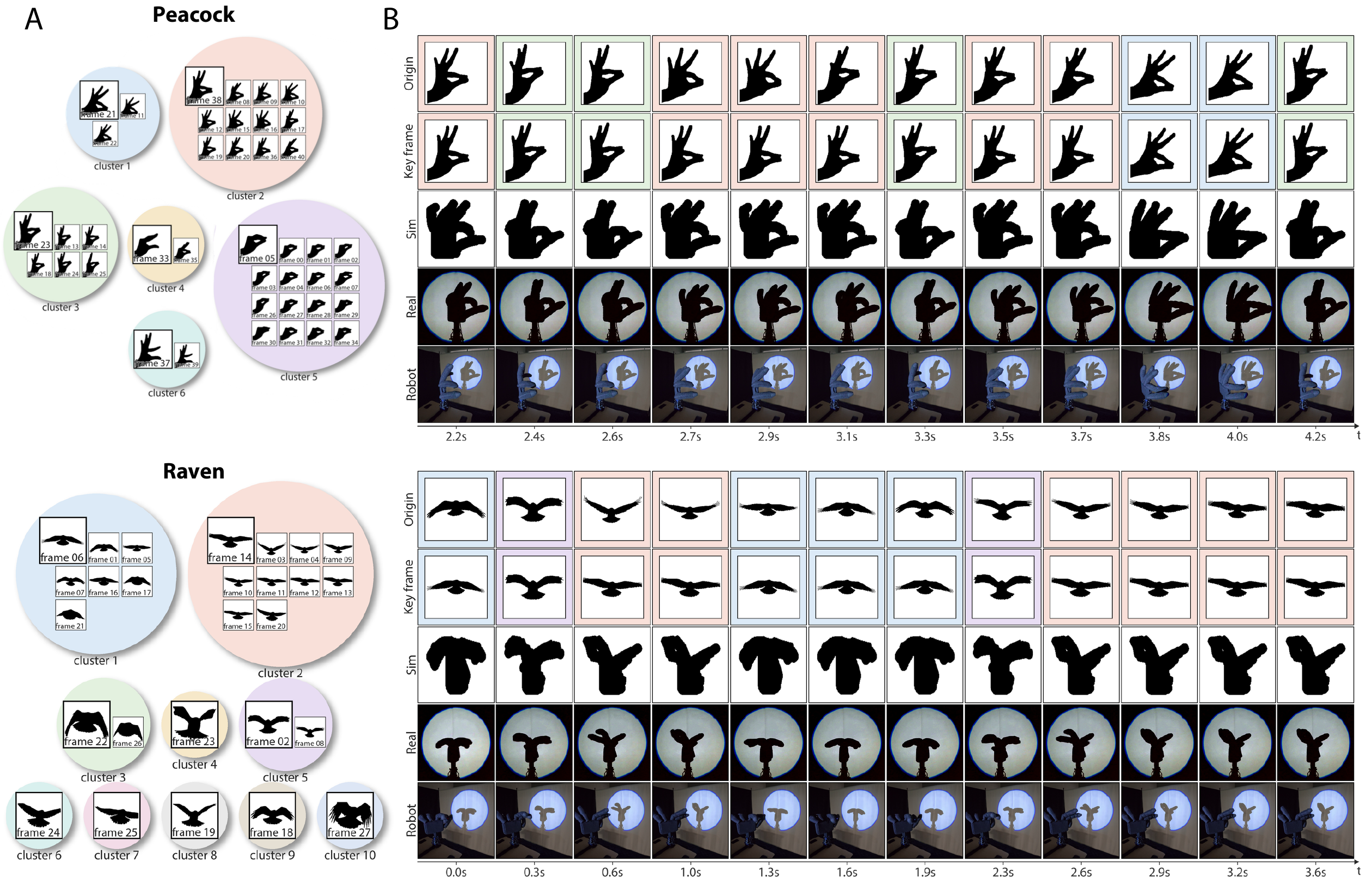}
    \caption{{$|$ \textbf{Keyframe extraction for video target imitation and sim-to-real transfer.} (\textbf{A}) Clustered frames from the peacock and raven video targets. The keyframe for each cluster is listed as the first image in the corresponding cluster. (\textbf{B}) Examples of 12 consecutive frames in both video targets. The top two rows demonstrate that replacing the original frames with their corresponding keyframes successfully preserves the motion style and expressive dynamics of the target videos. The bottom three rows show the simulation results and successful sim-to-real transfer for video imitation, where the transitions between consecutive frames are smooth.}}
    \label{fig:clustering_and_keyframe}
\end{figure}

However, we found that temporal regularization alone was insufficient when optimization was initialized independently for each frame. Because different random initializations frequently converged to different local optima, the temporal constraint often conflicted with the optimization process and produced limited improvements in transition quality. Instead, we introduced sequential optimization with inherited initialization. The first frame of a sequence was optimized using a larger search budget, and each subsequent frame was initialized from configurations sampled around the optimized solution of the previous frame (see ``Shadow Imitation Framework – Optimization’’ in Materials and Methods). This strategy constrained optimization to temporally consistent regions of the joint space before optimization began. A transition is considered successful only if both endpoint frames satisfy a reasonable global shadow performance with $\mathcal{L}_{\text{total}}\leq 0.35$, and the mean absolute joint change between adjacent frames satisfies $\mathrm{mean}(|\mathbf{\uptheta}_{i+1}-\mathbf{\uptheta}_i|)\leq 0.85$, and the maximum single-joint change satisfies $\max(|\mathbf{\uptheta}_{i+1}-\mathbf{\uptheta}_i|)\leq 2.00$. As shown in Table~\ref{tab:ablation_results}- Independent Per-frame Optimization, our previous method based on signal-frame optimization results in a low success rate. In contrast, our inherited initialization substantially improved transition success rates across all ablations. Combining expressive-region objectives, temporal regularization, and inherited initialization achieved the best performance, increasing the transition success rate from 34.5\% to 86.2\%. As shown in Figure~\ref{fig:qualitative}D, our final method achieves consistent shadow representations across frames, as indicated by the colored fingers appearing in similar regions of the images. In contrast, the baseline methods often use different finger and palm orientations to represent the shadow across frames, resulting in abrupt changes in finger color patterns.  

A third challenge arises from the redundancy present in long video sequences. Many consecutive frames contain nearly identical shadow shapes, making exhaustive optimization computationally inefficient. To reduce this redundancy, we clustered video frames using PCA feature representations and selected representative keyframes from each cluster (see ``Keyframe extraction for video targets’’ in Materials and Methods). We observed that frames within each cluster often corresponded to highly similar shadow configurations (fig.~\ref{fig:clustering_and_keyframe}A and fig.~\ref{fig:rest_of_key_frames}), suggesting that a sparse subset of frames could adequately represent the motion. Across the six video targets, including duck, deer, peacock, camel, wolf, and raven, the number of frames was reduced from 60, 16, 39, 12, 21, and 27 frames to 5, 5, 6, 3, 6, and 10 representative keyframes, respectively. These reductions corresponded to decreases of 91.7\%, 68.8\%, 84.6\%, 75.0\%, 71.4\%, and 63.0\% in the number of optimization targets. Despite this reduction, the keyframe representation preserved the characteristic motion style and expressive content of the original videos (fig.~\ref{fig:clustering_and_keyframe}B).

These components enabled dynamic communication through projected visual abstractions. Figure~\ref{fig:clustering_and_keyframe} shows representative results for peacock and raven imitation in both simulation and the physical robot. The robot successfully reproduced temporally coherent shadow motions while preserving important expressive structures and transferring the optimized trajectories from simulation to hardware. Across all six video targets, including hand-generated animal shadow puppetry and raw animal videos, the system generated recognizable and expressive shadow behaviors despite substantial embodiment differences between the target silhouettes and the robotic hand. Our full results are provided in the Supplementary Material, and our full video demonstration is provided in the Supplementary Movies.

\section*{Discussion}

We demonstrate that robots can communicate not only through their physical morphology, but also through projected visual abstractions generated by that morphology. By learning a shadow self-model, our 21-DoF robotic hand with compliant soft skin acquired a differentiable relationship between its internal joint configurations and the external silhouettes perceived by an observer. This capability allowed the robot to optimize its body not for direct physical interaction with the environment, but for visual expression through the appearance of an indirect representation of itself. In doing so, our work extends robotic expression beyond the body itself and introduces projected visual abstractions as a new medium for robotic communication.

A central insight of our work is that projected communication depends jointly on embodiment, self-modeling, and physical feasibility. The soft-skinned robotic hand addressed a hardware limitation of conventional rigid hands by reducing light leakage and enabling more continuous silhouettes, while retaining the kinematic controllability needed for optimization than fully soft hands. The learned shadow self-model addressed the many-to-one relationship between hand configurations and projected shadows by providing a differentiable forward model for shadow generation. The collision-aware refinement stage then reconciled the appearance objective with the constraints of physical embodiment, converting shadow-matching poses into executable robot configurations. These components allowed the robot to reproduce static hand gestures, hand-shadow puppetry, and animal motion targets in both simulation and the real world.

Dynamic shadow communication introduced additional challenges beyond static silhouette matching. Expressive information was often concentrated in small or moving regions, such as eyes, beaks, wing boundaries, and enclosed spaces. By incorporating expressive-region objectives, our robot better preserved these visually meaningful details. In addition, independent frame-wise optimization often produced discontinuous motions because different fingers could represent the same shadow region across adjacent frames. Sequential optimization with inherited initialization and temporal regularization improved motion consistency, while keyframe extraction reduced redundant optimization without eliminating the characteristic dynamics of the target shadows. Our results suggest that expressive robotic communication requires not only accurate target matching, but also careful considerations to the perceptual structure and temporal coherence of the generated behavior.

Existing expressive robotic systems primarily communicate through direct observation of their embodiment, including facial expressions, gestures, posture, locomotion, and artistic manipulation. In contrast, the communication channel explored in our work is external to the body itself. The shadows generated by our robot are not physical components of the robot, yet they remain tightly coupled to its embodiment and motion. Humans routinely communicate through similar transformed representations, including shadows, silhouettes, puppetry, drawings, reflections, and digital avatars. The ability to intentionally manipulate such representations may therefore expand the communicative bandwidth of robots beyond the physical limitations of their morphology and provide new opportunities for storytelling, entertainment, education, and human-robot interaction.

More broadly, shadow communication can be viewed as an extension of robotic self-modeling. Traditional self-models enable robots to reason about their own geometry, kinematics, dynamics, or physical state. Our shadow self-model predicts how the robot’s embodiment is transformed into an external visual representation perceived by an observer. This distinction shifts the objective from modeling the body itself to modeling the consequences of embodiment. Rather than asking ``What is my body doing?'', the robot must reason about ``What will an observer see if I do this?''. Such reasoning is conceptually related to perspective-taking and visual imagination, where an agent predicts how internal states manifest as external observations to others. Although our framework does not perform perspective-taking in the cognitive sense, it provides a computational mechanism for reasoning about how embodied actions can be transformed into observer-facing representations.

The shadow domain provides a particularly interesting testbed for studying this problem because it intentionally discards information. Multiple three-dimensional hand configurations can produce highly similar two-dimensional silhouettes, creating a fundamentally ambiguous inverse problem. As a result, successful shadow communication requires more than reproducing geometry. The robot must identify physically feasible embodiments that preserve information relevant to perception and interpretation in an abstract level. This challenge becomes even more pronounced during dynamic shadow performances, where expressive meaning is often conveyed through subtle motion patterns and visually salient details rather than solely through exact geometric accuracy. The expressive-region objectives and temporal optimization introduced in our work represent one approach toward addressing this broader problem of perceptually meaningful robotic communication.

Our results further highlight the importance of embodiment in enabling projected communication. Conventional rigid robotic hands are poorly suited for shadow performance because gaps between rigid linkages introduce light leakage and fragment the resulting silhouettes. Fully soft robotic hands, while capable of continuous contact, often sacrifice kinematic precision and controllability. Our hybrid rigid-soft design was motivated by the observation that human hand-shadow artists rely on both skeletal structure and compliant tissue to create expressive shadows. More generally, our findings suggest that embodiments optimized for physical manipulation may not necessarily be optimal for visual communication, and that future expressive robots in human environments may need to specifically consider designs around perceptual objectives beyond appearance.

The present framework has several limitations. First, the robot hand constrains the class of shadows that can be generated. Although the system successfully reproduced a diverse range of gestures, shadow puppets, and animal motions, targets with highly disconnected structures, extreme aspect ratios, or shapes far from hand morphology can still be challenging. One possible direction is to leverage more than one hand or external objects similar to human-shadow puppetry. Second, the current system relies on a controlled projection setup with fixed lighting and a fixed backdrop. More robust shadow communication will require adaptation to varying illumination conditions, projection geometries, and environmental surfaces. Third, optimization was performed offline and therefore does not yet support real-time interactive shadow generation. Though multiple aspects of our method has enabled efficient inference, more rapid search can enable real-time interactions.

We believe the broader implication of our work lies in enabling robots to reason about and manipulate transformed representations of themselves. Shadows represent only one example of a larger class of embodiment-dependent visual abstractions. Future robots may communicate through reflections, projected imagery, silhouettes, digital avatars, holographic displays, or other representations that are physically distinct from the robot body but causally linked to it. In this view, the body serves not only as a mechanism for acting on the world, but also as a generator of external symbols and representations. Developing robots capable of understanding and controlling such representations may open new directions for robotic communication, artistic expression, and embodied intelligence.

\section*{Materials and Methods}

\subsection*{Mechanical design}
\paragraph{Robot hand design for shadow imitation}

The inner components of the rigid hand skeleton were 3D printed using polylactic acid (PLA), and the outer soft skin was fabricated from foaming thermoplastic polyurethane (TPU). Foaming TPU undergoes a thermally activated expansion process, which produces a foam-like structure with high compliance. Our material selection provides both a light-blocking, sealed surface for shadow formation and sufficient softness to conform to the motion of the underlying skeleton while allowing reasonable in-between finger contacts. The cover was printed at a nozzle temperature of $270^{\circ}\mathrm{C}$ with a $12\%$ gyroid infill. A wall-less printing setting was used to further reduce bending stiffness and improve flexibility. The expanded TPU structure and internal gyroid lattice maintained sufficient opacity to block light during shadow projection.

Our robotic hand is actuated by 21 LX-15D serial bus servos. Each finger has an identical kinematic structure with four degrees of freedom (DoF), and an additional 1-DoF wrist joint is located at the base. The specific dimension is shown in Figure~\ref{fig:dimension}. To improve mechanical stability at the wrist, we incorporate a metal rotary platform along with additional 3D-printed clamps for reinforcement. The controller operates at a frequency of 40\,Hz. Motor commands are dispatched in asynchronous batches (seven motors per batch) with a 10\,ms delay between batches. 

\subsection*{Shadow imitation framework}

\paragraph{Model architecture}
Our shadow self-model is composed of two components: an analytical forward kinematics (FK) module followed by a neural network. The neural network uses two fully connected layers to project the flattened FK transformation matrices into a low-resolution feature map, followed by a five-layer up-convolutional decoder that generates the shadow image.

The FK module maps a 21-dimensional joint configuration $\boldsymbol{\uptheta} \in \mathbb{R}^{21}$ to the spatial poses of all joints in the kinematic chain. Specifically, it computes a set of $4 \times 4$ homogeneous transformation matrices that encode both rotation and translation, providing a complete representation in $SE(3)$. This formulation introduces strong geometric priors of our robotic hand while preserving differentiability, enabling smoother end-to-end gradient backpropagation through the entire computational graph.

For each joint $i$, the local transformation is defined as
\begin{equation}
\mathbf{T}_i(\theta_i) =
\begin{bmatrix}
\mathbf{R}_i(\theta_i) & \mathbf{t}_i \\
\mathbf{0}^\top & 1
\end{bmatrix},
\end{equation}
where $\mathbf{R}_i(\theta_i) \in SO(3)$ is the rotation matrix parameterized by the joint angle $\theta_i$, and $\mathbf{t}_i \in \mathbb{R}^3$ is the fixed translation defined by the kinematic structure.

The rotation matrix is computed using Rodrigues' formula:
\begin{equation}
\mathbf{R}_i(\theta_i) = \mathbf{I} + (\sin \theta_i)\mathbf{K}_i + (1 - \cos \theta_i)\mathbf{K}_i^2,
\end{equation}
where $\mathbf{K}_i$ is the skew-symmetric matrix corresponding to the unit rotation axis of joint $i$.

The global transformation of each joint is obtained by recursively composing transformations along the kinematic tree:
\begin{equation}
\mathbf{T}_i^{\mathrm{glob}} =
\begin{cases}
\mathbf{T}_i(\theta_i), & \text{if } i \text{ is the root}, \\
\mathbf{T}_{p(i)}^{\mathrm{glob}} \, \mathbf{T}_i(\theta_i), & \text{otherwise},
\end{cases}
\end{equation}
where $p(i)$ denotes the parent of joint $i$.

All operations are differentiable, allowing gradients to propagate through both rotational and translational components of the kinematic chain. This enables the model to learn spatially consistent joint configurations while respecting the underlying hand geometry.

Given a 21-dimensional joint-angle vector $\mathbf{q}\in\mathbb{R}^{21}$, the forward model first applies a differentiable forward-kinematics layer built from the Unified Robot Description Format (URDF) of our robotic hand. For each joint in the predefined kinematic order, the layer composes the local joint rotation with the static parent-child transform and the selected link/frame offset, producing base-frame global transformations
$\mathbf{T}(\mathbf{q})=\{\mathbf{T}_i^{\mathrm{glob}}\}_{i=1}^{21}\in\mathbb{R}^{21\times4\times4}$.
The transformation tensor is flattened into a 336-dimensional feature vector and passed to a ResNet-style shadow decoder. The decoder maps this vector through fully connected layers to a $128\times8\times8$ latent feature map, applies residual blocks, and progressively upsamples it with transposed convolutions to generate a single-channel shadow image
$\mathbf{I}\in[0,1]^{1\times256\times256}$.

The resulting set of transformations $\{\mathbf{T}_i^{\mathrm{glob}}\}_{i=1}^{21}$ is stacked into a tensor $\mathbf{T} \in \mathbb{R}^{21 \times 4 \times 4}$, flattened into a feature vector $\mathbf{z} \in \mathbb{R}^{336}$, and fed into the shadow image decoder. The decoder outputs a shadow image $\mathbf{I} \in \mathbb{R}^{256 \times 256}$. 

The baseline inverse shadow self-model $InvS_{\psi}$ follows an encoder-style architecture that maps a shadow image back to the corresponding joint configuration. Given a shadow image $\mathbf{I} \in \mathbb{R}^{1 \times 256 \times 256}$, the model applies five stride-2 convolutional downsampling blocks with channel dimensions $16$, $32$, $64$, $128$, and $128$. Each convolutional block is followed by batch normalization and ReLU activation, and residual blocks are inserted between downsampling stages to improve feature representation.

After the final downsampling stage, the feature map is processed by two additional residual bottleneck blocks. The resulting latent feature is flattened and passed through a fully connected regression head, consisting of a $1024$-dimensional hidden layer followed by a linear output layer. The final output is a 21-dimensional joint vector:
\begin{equation}
\hat{\boldsymbol{\uptheta}} = InvS_{\psi}(\mathbf{I}), 
\quad \hat{\boldsymbol{\uptheta}} \in \mathbb{R}^{21}.
\end{equation}

\paragraph{Data collection} We collected 9,249,784 pairs of joint configurations and corresponding shadow images in simulation. Among the collected data, $75\%$ of the samples were generated through random sampling under predefined joint constraints. The wrist rotation was uniformly sampled from $[-90^\circ, 90^\circ]$. The MCP joints were sampled within finger-specific ranges, including $[0^\circ, 90^\circ]$ for the thumb, $[0^\circ, 60^\circ]$ for the ring and pinky fingers, and $[-60^\circ, 0^\circ]$ for the index and middle fingers. The remaining finger joints were sampled from approximately $[-90^\circ, 90^\circ]$.

To further enrich the dataset with sparse finger motions, $10\%$ of the samples were generated using single-finger configurations, while $15\%$ were generated using two-finger configurations. In these structured poses, either one active finger or two active fingers were selected, while the remaining fingers stayed near a resting pose. These structured single- and double-finger samples were introduced to augment the dataset and improve the representation of isolated finger motions.

\paragraph{Training} We collected $N$ pairs of joint configurations and corresponding shadow images, denoted as $\{(\mathbf{\uptheta}_i, \mathbf{I}_i)\}_{i=1}^{N}$, in simulation. The dataset was divided into training, validation, and testing sets with a ratio of $0.7$, $0.2$, and $0.1$, respectively. Given a joint configuration $\bf{\uptheta}_i$, we trained the shadow self-model $S_{\phi}(\mathbf{\uptheta})$, parameterized by $\phi$, to predict the corresponding shadow image $\hat{\mathbf{I}}_i = S_{\phi}(\boldsymbol{\uptheta}_i)$. The model was optimized with an $\mathcal{L}_{\text{MAE}}(\hat{\mathbf{I}},\mathbf{I})$ reconstruction loss:
\begin{equation}
\min_{\phi} \sum_{i=1}^{N} \left\| \hat{\mathbf{I}}_i - \mathbf{I}_i \right\|_1, 
\quad \text{where} \quad 
\hat{\mathbf{I}}_i = S_{\phi}(\boldsymbol{\uptheta}_i).
\end{equation}

In the baseline inverse model training, we used the same dataset. Given a shadow image $\mathbf{I}_i$, we trained the inverse shadow self-model $InvS_{\psi}(\mathbf{\uptheta})$, parameterized by $\psi$, to predict the corresponding joint configuration $\hat{\mathbf{\uptheta}}_i = InvS_{\psi}(\mathbf{I}_i)$. The model was optimized using an $\mathcal{L}_{\text{MSE}}(\hat{\mathbf{\uptheta}},\mathbf{\uptheta})$ reconstruction loss:
\begin{equation}
\min_{\psi} \sum_{i=1}^{N} \left\| \hat{\mathbf{\uptheta}}_i - \mathbf{\uptheta}_i \right\|_1, 
\quad \text{where} \quad 
\hat{\mathbf{\uptheta}}_i = S_{\psi}(\mathbf{I}_i).
\end{equation}

\paragraph{Optimization}
To make the image loss focus on the informative shadow region, we first cropped both the target and predicted shadow images using the bounding box of the shadow region. The bounding box was computed from pixels darker than 0.5, padded by five pixels on each side within the image boundary, and resized to a common resolution of $256 \times 256$. All image-space losses were then computed on these aligned shadow crops rather than on the full canvas.

For independent per-frame optimization, we optimized each frame separately with a batch size of $256$ for $2000$ iterations. For sequential optimization with inherited initialization, we used a batch size of $500$ and $3000$ iterations for the first frame, and then optimized later frames with a batch size of $128$ and $1200$ iterations using the previous optimized frame as initialization. For inherited initialization, Gaussian noise with standard deviation $0.05$ radians was added to the previous optimized joint configuration.

For refinement, we used a hill-climbing search initialized from the best joint configuration predicted by the previous stage. The search consisted of 8 runs with different initial noise samples, and each run was executed for 1000 steps. Restart initialization was generated by adding Gaussian noise with standard deviation $0.25$ radians around the initial pose. During local search, joint perturbations were sampled with an initial standard deviation of $0.08$ radians, clipped to the joint limits, and accepted only when they reduced the weighted image loss. The perturbation scale was multiplied by $0.995$ after each step and lower-bounded by $0.01$ radians. The gradient-based neural optimization was conducted on a single NVIDIA L40S GPU, whereas the hill-climbing search, including simulation-based rendering and local pose perturbation, ran on a 12th Gen Intel(R) Core(TM) i9-12900K CPU.

In all the optimizations, the weights of the base loss were set to $\lambda_{\text{MAE}} = 0.06$, $\lambda_{\text{IoU}} = 0.4$, and $\lambda_{\text{CLIP}} = 0.02$. The weights for the expressive-region and temporal consistency objectives were set to $\lambda_{\text{Exp\_IoU}} = 1.0$, $\lambda_{\text{Exp\_CLIP}} = 0.05$, and $\lambda_{\text{temp}} = 0.2$ when applied during optimization.

\subsection*{Expressive region}

The expressive region is a binary mask that covers the moving portion and the enclosed empty portion of the shadow. Given a sequence of target frames from the video, we first binarized the target images by thresholding each frame to separate the shadow from the background. The expressive region was then computed from the resulting binary shadow masks. Pixels that remain part of the object in every frame were identified as the stationary region. For a given frame, the expressive region consisted of the object pixels after removing this stationary region.

Additionally, any empty holes completely enclosed within the shadow were included in the expressive region. To identify such holes, we inverted the object mask and computed its connected components. Any connected component in the inverted image that did not touch the image border was considered a fully enclosed hole and was added to the expressive region mask. 

\subsection*{Keyframe extraction for video
targets}

For a given video target, we first sampled frames at 3\,Hz, 3.3\,Hz, 5.48\,Hz, 3.44\,Hz for the hand shadows of duck, deer, peacock and camel respectively, and used 3.1 Hz for the real raven,  2.35 Hz for the real wolf example. For RGB image shadow targets performed by human hand with pixel values ranging from $0$ to $255$, each frame was binarized into a silhouette using a predefined intensity threshold of $128$. For general video targets, we used Segment Anything~\cite{kirillov2023segment} to extract the object mask, which was then converted into a binary silhouette image. Optimizing a smooth and continuous joint trajectory became increasingly challenging and slow as the number of frames grew. To address this, we proposed a grouping strategy based on frame similarity. Specifically, we clustered frames in the target video and selected representative keyframes for optimization. This reduced computational complexity while preserving the expressiveness and temporal coherence of the motion. Therefore, in our video target imitation, instead of optimizing over the full sequence, we optimized over only the selected keyframes. During imitation, each frame was assigned the optimized joint configuration of its corresponding representative keyframe based on the clustering.

To identify keyframes, we developed an iterative clustering method based on principal component analysis (PCA) of the expressive regions in each frame. Given $N$ frames, each expressive region mask was resized to $64 \times 64$ pixels, flattened into a $4096$-dimensional vector, and normalized to the range $[0,1]$. These vectors were then projected onto the top $50$ principal components using PCA. Pairwise $\ell_2$ distances were computed between all PCA embeddings and normalized by the maximum distance, resulting in a distance matrix $D \in \mathbb{R}^{N \times N}$, where smaller values indicated higher similarity.

We then performed iterative agglomerative clustering by repeatedly merging the pair of clusters with the smallest distance. After each merge, cluster features were updated by averaging the embeddings of their member frames. After the first merge, the number of clusters was reduced to $N-1$, and the distance matrix was updated accordingly to $D \in \mathbb{R}^{(N-1) \times (N-1)}$. This process continued until a similarity threshold $\tau \in [0,1]$ was reached. A larger $\tau$ enforced stricter similarity and typically resulted in a greater number of clusters.

The threshold $\tau$ thus controlled the distinctiveness of clusters. Once clustering was complete, one representative frame was selected from each cluster. The representative frame was chosen as the one whose embedding had the smallest $\ell_2$ distance to the average of the PCA embeddings. These representative frames formed the final targets used for generating the video shadow imitation.

\subsection*{Motion planning for sim-to-real transfer}
Given a sequence of optimized joint configurations, we developed a simple motion planning pipeline in simulation and directly deployed the resulting collision-free trajectory to the real robot. Our interpolation scheme used a three-pass strategy to produce smooth and physically valid motion between key poses.

First, we fit a global cubic spline with not-a-knot oundary conditions through the full keyframe sequence prepended with the home configuration, ensuring $C^{2}$-continuous motion across the entire trajectory. When a cubic spline could not be constructed, the system fell back to per-segment
linear interpolation. The interpolated trajectory was then evaluated in simulation with self-collision enabled, and each proposed pose was scored using the collision metric $S_{\text{col}}$ defined in
the following paragraph, where configurations with no contact receive $S_{\text{col}}=0$. When the score exceeded the safety threshold, the script performed noise-based recovery. It sampled $50$ random $\pm 5^{\circ}$ perturbations around the previously accepted pose, scored each candidate by the collision cost along its path to the current spline target, kept the three lowest-scoring candidates, and broke ties with a one-step lookahead to the next spline target. The last frame of every segment was always the exact keyframe and was never perturbed. A lightweight single-pose collision check was also used during fingertip refinement to quickly reject already-colliding frames
or unsafe fingertip-only updates. In the final pass, we adjusted only the five distal (TIP) joints. Because the noise-recovery search was more constrained at the proximal joints, perturbing them risked
self-collisions across large portions of the hand. The remaining deviation from the desired keyframe pose was naturally most pronounced at the fingertips, which also dominated the silhouette of the projected shadow. We defined this residual as the squared $\ell_{2}$ distance, taken over the five tip joints, between the post-recovery joint angles and those of the nearest keyframe. For each non-keyframe frame, the refinement stage generated $20$ biased candidates, each a $5\%$--$25\%$
step toward the nearest keyframe's tip pose plus $\pm 2^{\circ}$ jitter, and accepted a candidate only when it was both collision-free and strictly reduced this residual, keeping the rest of the hand frozen so the collision safety established in the first pass was preserved.

The collision score was defined as
\[
S_{\mathrm{col}} =
w_{\mathrm{pen}}
\left(
\sum_i \max(0,-d_i)
+
\max(0, p_{\mathrm{soft}} - d_{\min})
\right)
+
w_{\mathrm{nf}}
\max(0, F_{\max} - F_{\mathrm{soft}}),
\]
where \(d_i\) was the simulation contact distance at contact \(i\), \(d_{\min}=\min_i d_i\), \(F_{\max}=\max_i F_i\), \(w_{\mathrm{pen}}=500\), and \(w_{\mathrm{nf}}=0.2\). Negative \(d_i\) values indicated penetration, so the score penalized total penetration, penetration beyond the soft threshold \(p_{\mathrm{soft}}\), and normal force beyond \(F_{\mathrm{soft}}\). Configurations with no contact received \(S_{\mathrm{col}}=0\).


\clearpage 

%
\bibliography{science_template} 
\bibliographystyle{sciencemag}

%
%
%
%
%
%


\noindent \textbf{Acknowledgments}: We thank Jacob Lee for the initial version of the code development for the motor controller. \textbf{Funding}: This work is supported by DARPA TIAMAT program under award HR00112490419, ARO under award W911NF2410405, NSF ERC PreMiEr under award 2133504, and ARL STRONG program under awards W911NF2320182, W911NF2220113, and W911NF242021. \textbf{Author contributions}: B.C. and J.L. conceived and designed the research. D.Y., B.W., and J.L. designed simulations and performed physical experiments. All authors analyzed data and wrote the manuscript. \textbf{Competing interests}: The authors declare no competing interests. \textbf{Data, code and materials availability}: The code is available at: https://github.com/generalroboticslab/shadow-Xdance.
The high-resolution version of all the supplementary videos can be accessed at  \url{https://figshare.com/s/a0a573cc51a330cc1b32}. Materials were commercially available.


\subsection*{Supplementary materials}
Figs. S1 to S18\\
Movie 1 \\
Movie S1 to S7\\

\newpage

\clearpage


\renewcommand{\thefigure}{S\arabic{figure}}
\renewcommand{\thetable}{S\arabic{table}}
\renewcommand{\theequation}{S\arabic{equation}}
\renewcommand{\thepage}{S\arabic{page}}
\setcounter{figure}{0}
\setcounter{table}{0}
\setcounter{equation}{0}
\setcounter{page}{1} 


\begin{center}
\section*{Supplementary Materials for\\ \scititle}

\author{
	Danyang Yan$^{1\dagger}$,
	Boyuan Wang$^{2\dagger}$,
    Jiaxun Liu$^{2\dagger}$,
    Boyuan Chen$^{1,2,3\ast}$\\
    \normalsize{$^{1}$Department of Electrical and Computer Engineering, Duke University}\\
    \normalsize{$^{2}$Department of Mechanical Engineering and Materials Science, Duke University}\\
    \normalsize{$^{3}$Department of Computer Science, Duke University}\\
    \normalsize{$^\ast$To whom correspondence should be addressed; E-mail: boyuan.chen@duke.edu.}\\
    \normalsize{$^\dagger$These authors contributed equally to this work.}
}
\end{center}

\subsubsection*{This PDF file includes:}
Supplementary Text \\
Figures S1 to 18\\
Captions for Movies S1 to S7\\

\subsubsection*{Other Supplementary Materials for this manuscript:}
Movie 1 \\
Movies S1 to S7\\

\newpage


\subsection*{Full imitation targets and qualitative results}
We present the complete qualitative demonstration results by comparing the target image, projected shadow, and real-world robot configuration for all 26 American Sign Language(ASL) letters. For the selected keyframes from each video example, we also compare the projected shadows in RGB and binary formats, as well as the extracted expressive regions. The target shadows are first extracted using simple thresholding or Segment Anything from the RGB image, and then binarized.

\begin{figure}[hbt!]
    \centering
    \includegraphics[width=\textwidth]{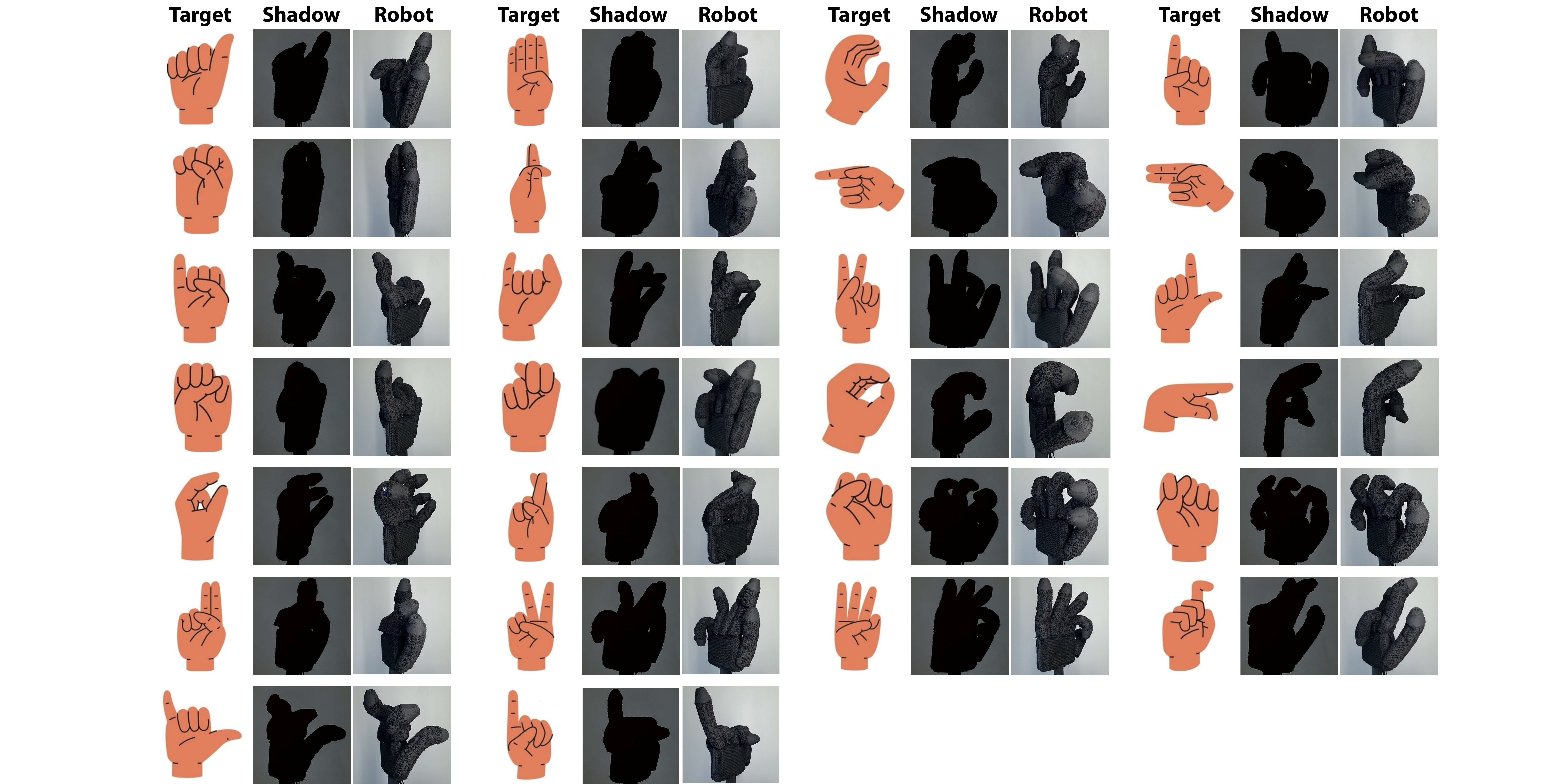}
    \caption{\textbar{} \textbf{Qualitative results of shadow-based self-modeling for the full ASL alphabet demonstration.}}
    \label{fig:full_alphabet}
\end{figure}

\begin{figure}
    \centering
    \includegraphics[width=\textwidth]{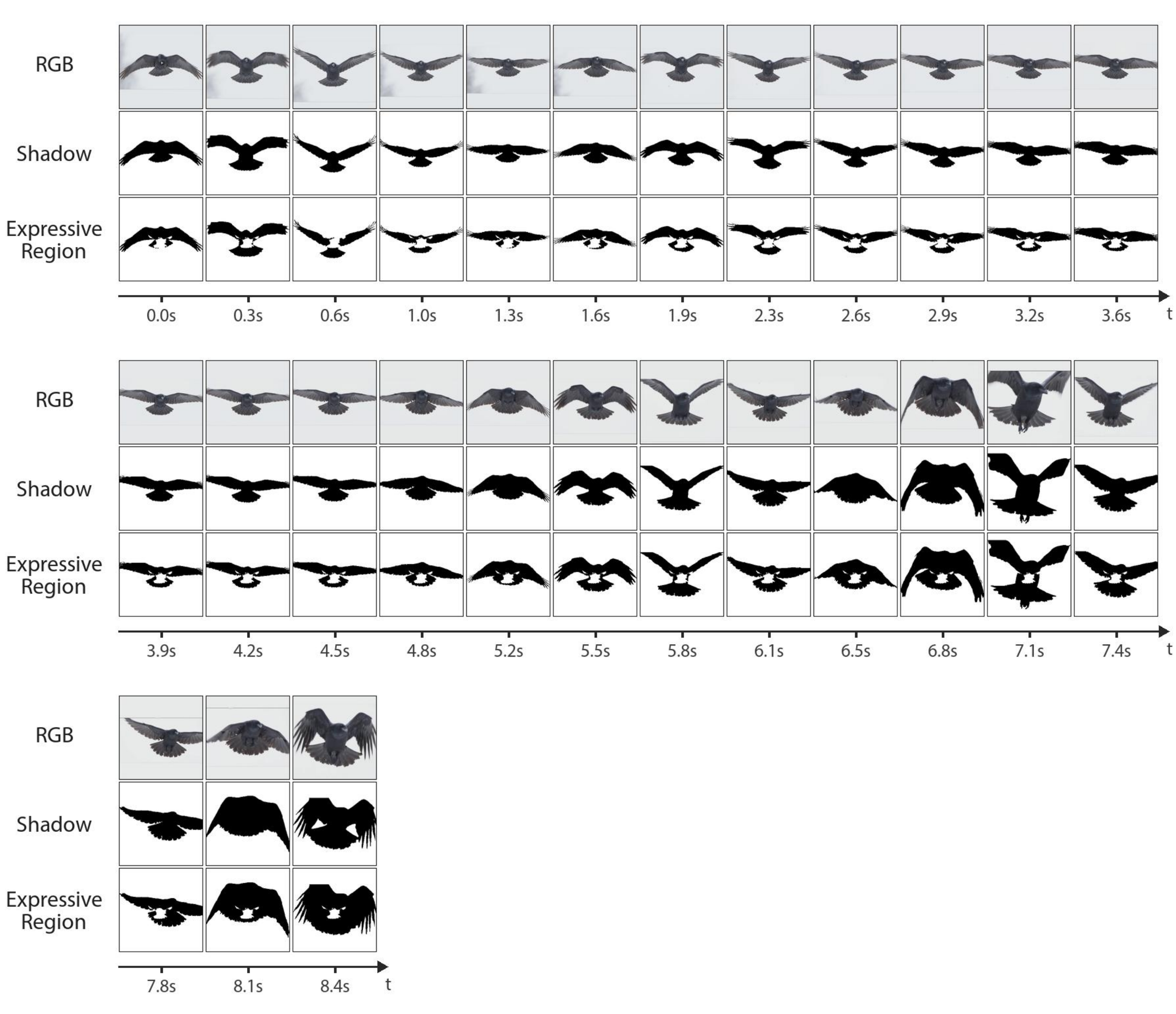}
    \caption{\textbar{} \textbf{Qualitative results of shadow-based self-modeling for the real raven video example. All 27 sampled frames are shown.}}
    \label{fig:full_bird_target}
\end{figure}

\begin{figure}
    \centering
    \includegraphics[width=\textwidth]{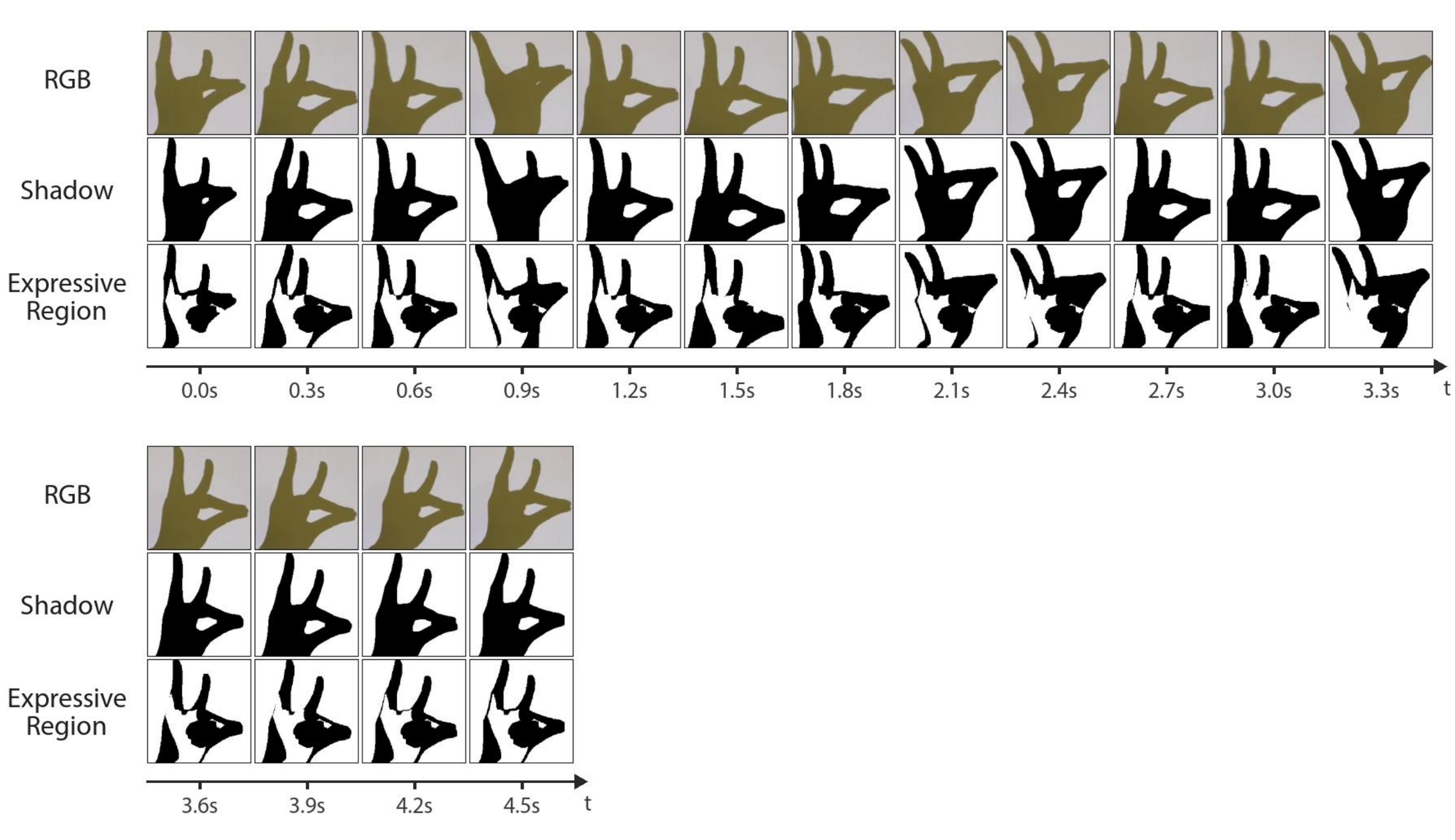}
    \caption{\textbar{} \textbf{Qualitative results of shadow-based self-modeling for the deer video example. All 16 sampled frames are shown.}}
    \label{fig:full_deer}
\end{figure}

\begin{figure}
    \centering
    \includegraphics[width=0.6\textwidth]{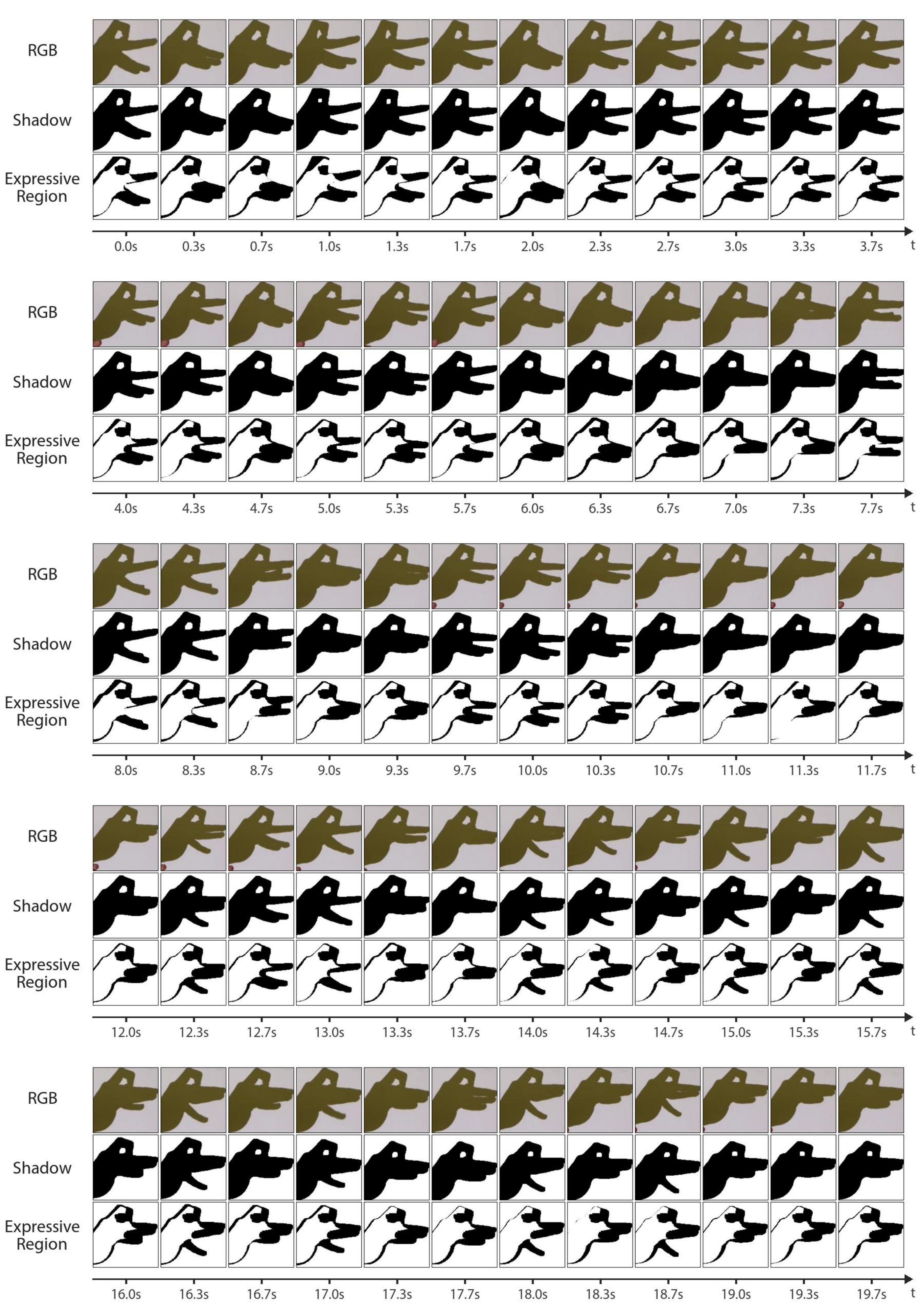}
    \caption{\textbar{} \textbf{Qualitative results of shadow-based self-modeling for the duck video example. All 60 sampled frames are shown.}}
    \label{fig:full_duck}
\end{figure}

\begin{figure}
    \centering
    \includegraphics[width=\textwidth]{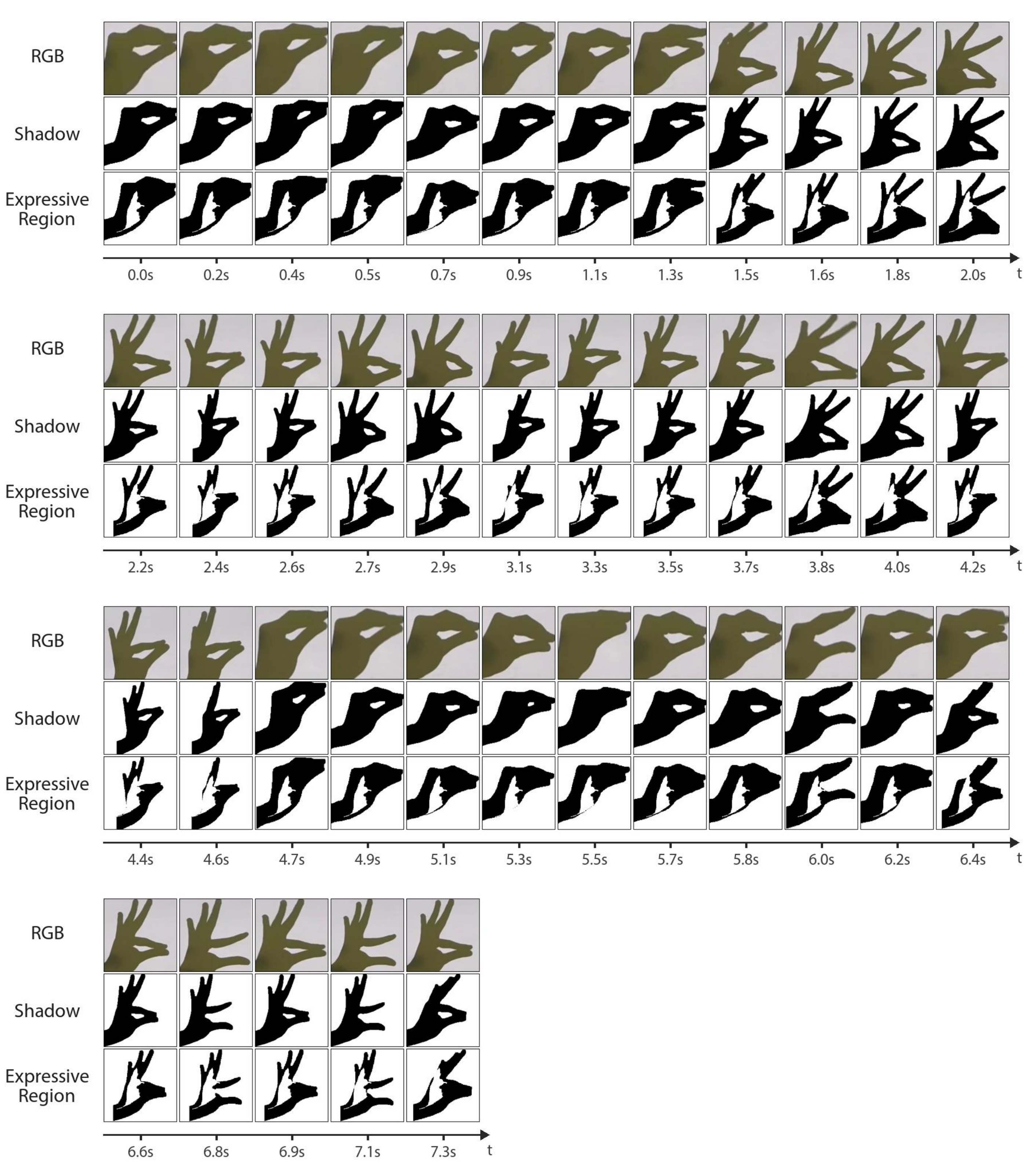}
    \caption{\textbar{} \textbf{Qualitative results of shadow-based self-modeling for the peacock video example. All 41 sampled frames are shown.}}
    \label{fig:full_peacock2}
\end{figure}

\begin{figure}
    \centering
    \includegraphics[width=\textwidth]{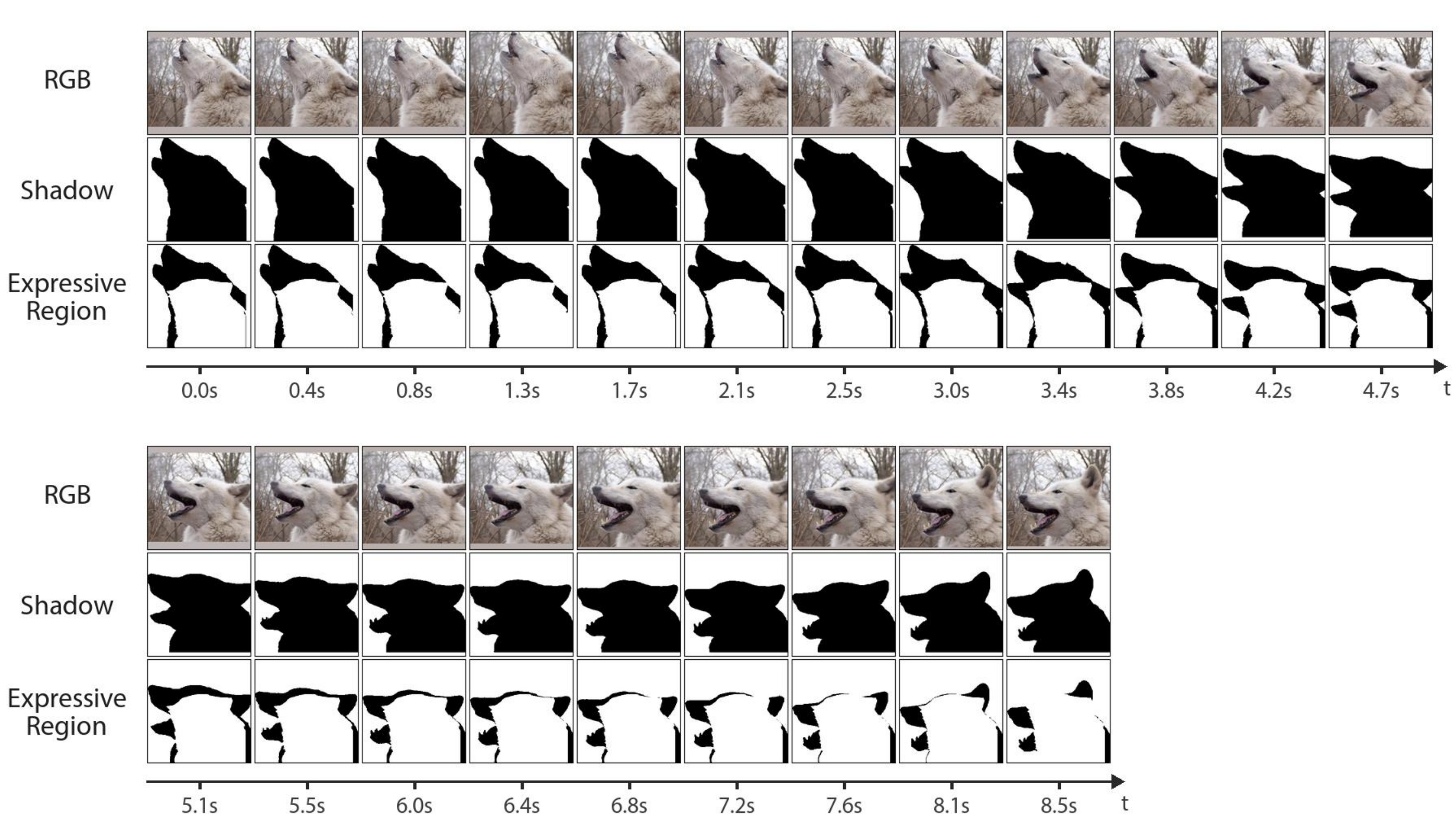}
    \caption{\textbar{} \textbf{Qualitative results of shadow-based self-modeling for the real wolf video example. All 20 sampled frames are shown.}}
    \label{fig:full_wolf}
\end{figure}

\begin{figure}
    \centering
    \includegraphics[width=0.8\textwidth]{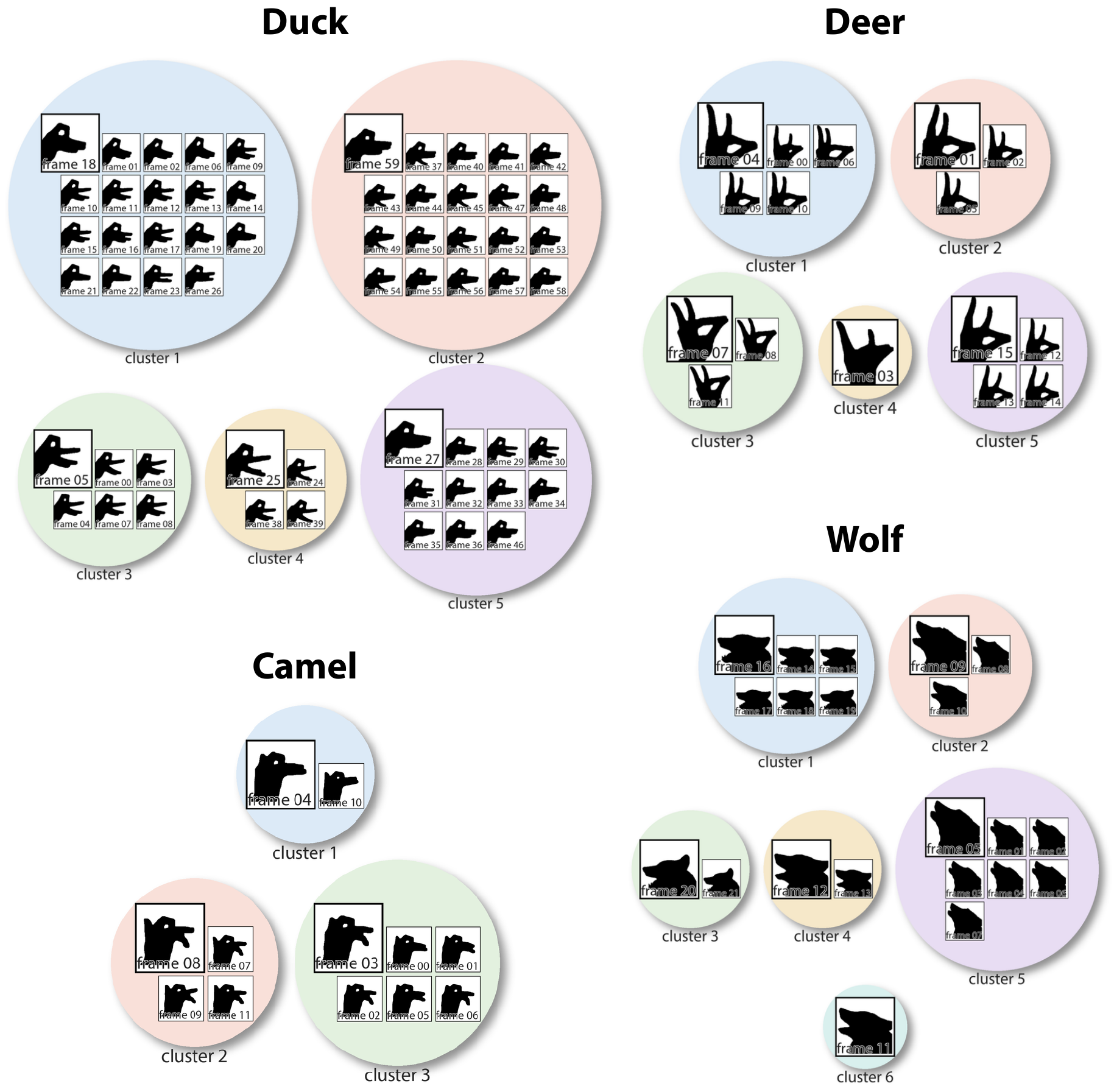}
    \caption{\textbar{} \textbf{Clustering results for the duck, deer, wolf, and peacock video examples.} Each colored circle denotes one cluster, and the enlarged image within each cluster represents the selected representative frame.}
    \label{fig:rest_of_key_frames}
\end{figure}

The ASL alphabet results are shown in Figure~\ref{fig:full_alphabet}, and the corresponding animal keyframe results are shown in Figures~\ref{fig:full_bird_target}--\ref{fig:full_wolf}. Figure~\ref{fig:rest_of_key_frames} presents the clustering results for the duck, deer, wolf, and peacock demos. Each colored circle represents one cluster, and the enlarged image within each cluster shows the representative frame.

The reproduced video frames, generated by replacing the original frames with their corresponding representative frames, are shown in Figures~\ref{fig:full_bird_keyframe_imitation}--~\ref{fig:full_wolf_keyframe_imitation}. Each figure compares the binarized shadow, representative keyframe from clustering, optimized shadow from the simulator, real-world projected shadow, and corresponding real robot configuration.

\clearpage

\begin{figure}
    \centering
    \includegraphics[width=0.8\textwidth]{figure/final/fig_S8.pdf}
    \caption{\textbar{} \textbf{Qualitative results of shadow-based self-modeling for real raven video demonstration.} The figure compares the original shadow, binarized representative keyframe, optimized simulation shadow, real-world projected shadow, and corresponding real robot configuration.}
    \label{fig:full_bird_keyframe_imitation}
\end{figure}

\begin{figure}
    \centering
    \includegraphics[width=\textwidth]{figure/final/fig_S9.pdf}
    \caption{\textbar{} \textbf{Qualitative results of shadow-based self-modeling for the camel video demonstration.} The figure compares the original shadow, binarized representative keyframe, optimized simulation shadow, real-world projected shadow, and corresponding real robot configuration.}
    \label{fig:full_camel_keyframe_imitation}
\end{figure}

\begin{figure}
    \centering
    \includegraphics[width=\textwidth]{figure/final/fig_S10.pdf}
    \caption{\textbar{} \textbf{Qualitative results of shadow-based self-modeling for the deer video demonstration.} The figure compares the original shadow, binarized representative keyframe, optimized simulation shadow, real-world projected shadow, and corresponding real robot configuration.}
    \label{fig:full_deer_keyframe_imitation}
\end{figure}

\begin{figure}
    \centering
    \includegraphics[width=0.5\textwidth]{figure/final/fig_S11.pdf}
    \caption{\textbar{} \textbf{Qualitative results of shadow-based self-modeling for the duck video demonstration.} The figure compares the original shadow, binarized representative keyframe, optimized simulation shadow, real-world projected shadow, and corresponding real robot configuration.}
    \label{fig:full_duck_keyframe_imitation}
\end{figure}

\begin{figure}
    \centering
    \includegraphics[width=0.7\textwidth]{figure/final/fig_S12.pdf}
    \caption{\textbar{} \textbf{Qualitative results of shadow-based self-modeling for the peacock video demonstration.} The figure compares the original shadow, binarized representative keyframe, optimized simulation shadow, real-world projected shadow, and corresponding real robot configuration.}
    \label{fig:full_peacock_keyframe_imitation}
\end{figure}

\begin{figure}
    \centering
    \includegraphics[width=\textwidth]{figure/final/fig_S13.pdf}
    \caption{\textbar{} \textbf{Qualitative results of shadow-based self-modeling for real wolf video demonstration.} The figure compares the original shadow, binarized representative keyframe, optimized simulation shadow, real-world projected shadow, and corresponding real robot configuration.}
    \label{fig:full_wolf_keyframe_imitation}
\end{figure}

\begin{figure}
    \centering
    \includegraphics[width=1\textwidth]{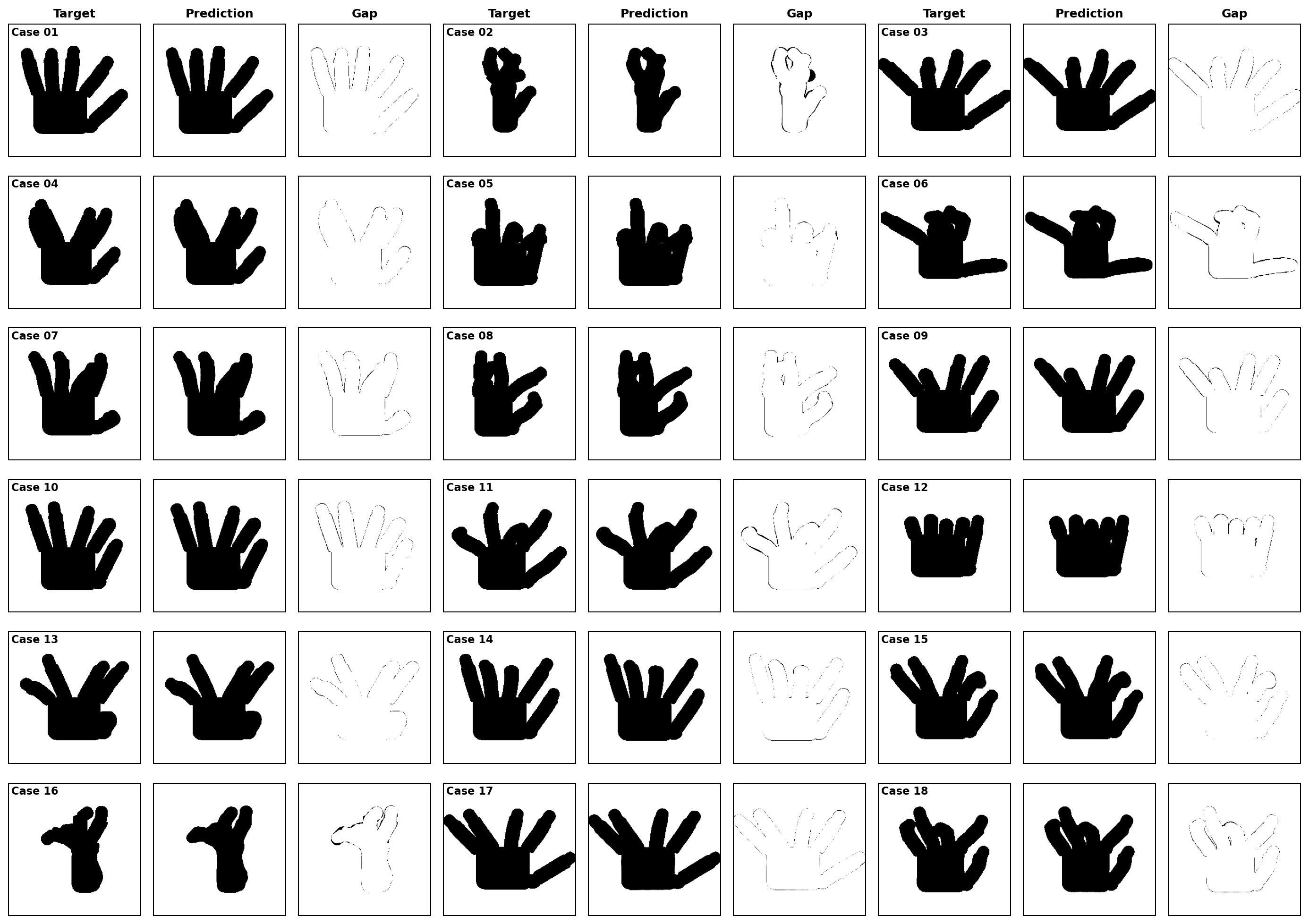}
    \caption{{$|$ \textbf{Qualitative results for shadow self-model} }}
    \label{fig：shadow self-model qualitative}
\end{figure}

\begin{figure}[t!]
	\centering
	\includegraphics[width=1\textwidth]{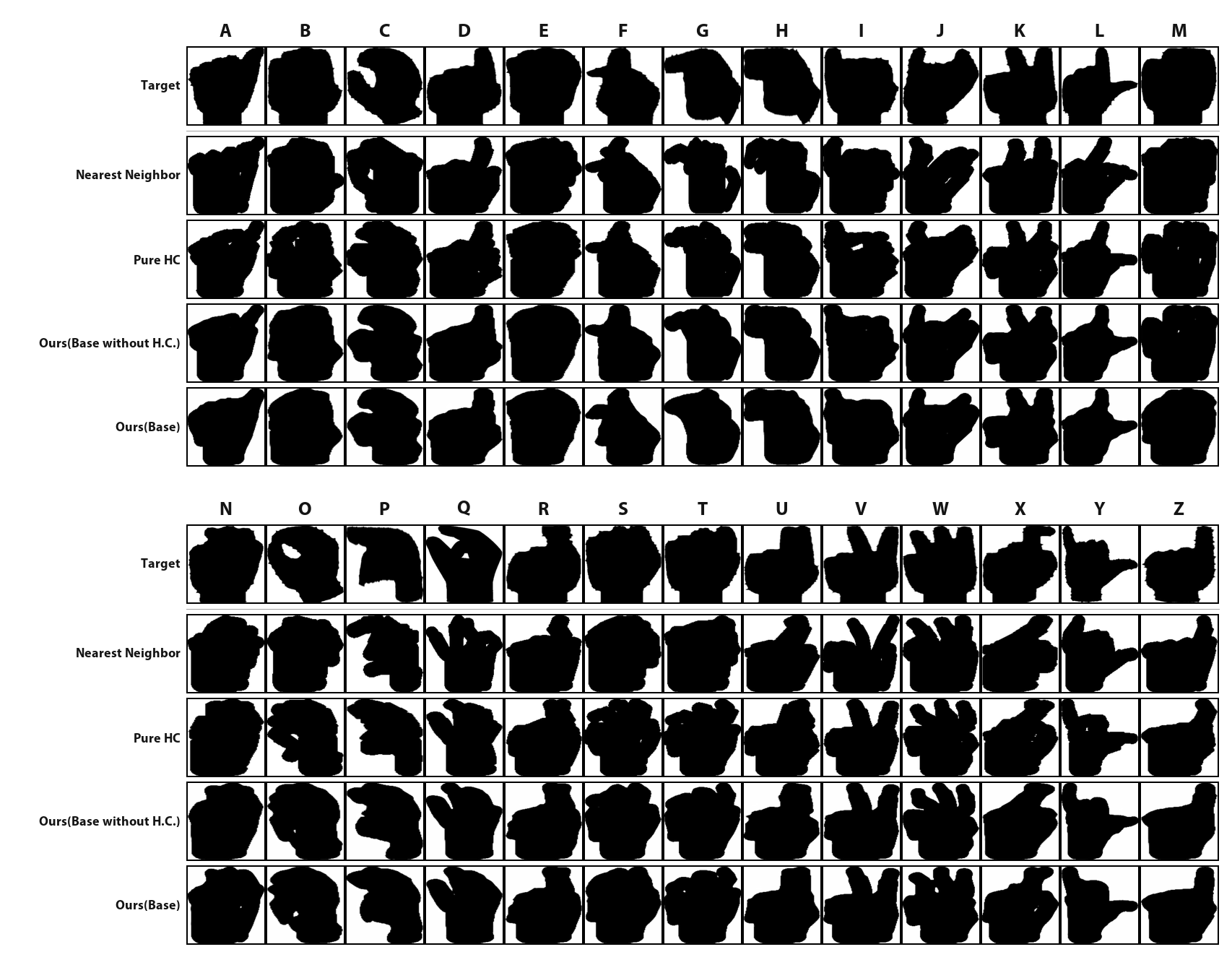} 
	\caption{\textbf{Qualitative ablation results for 26 ASL targets.}}
	\label{fig:paper_alphabet_qualitative}
\end{figure}

\begin{figure}[t!]
	\centering
	\includegraphics[width=1\textwidth]{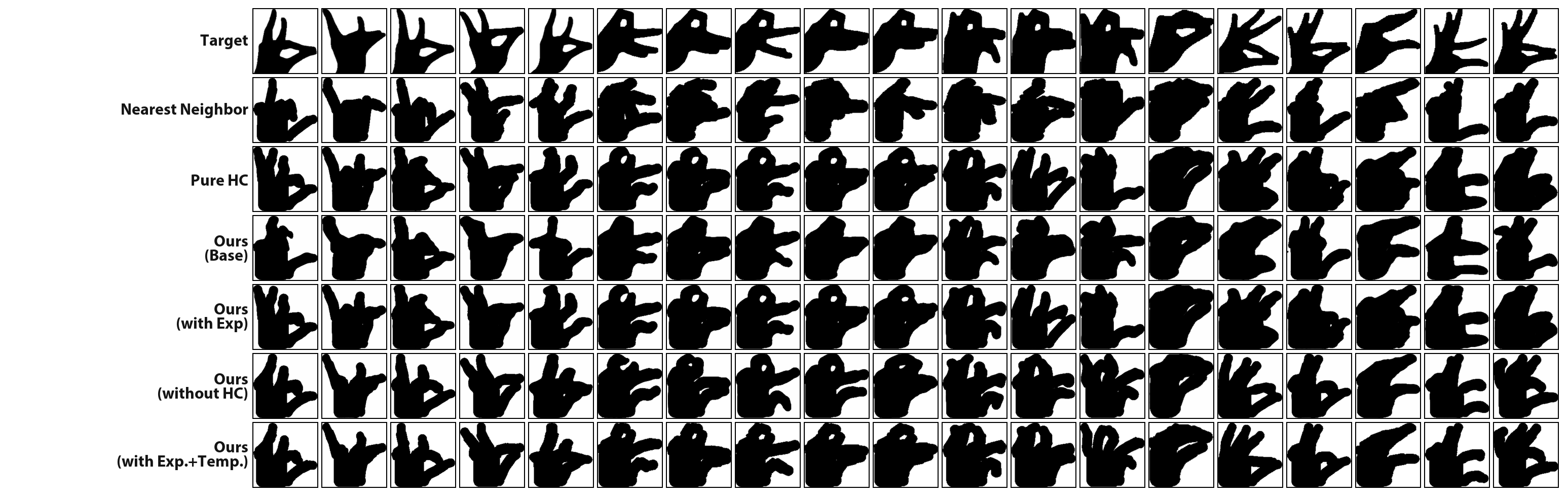} 
	\caption{\textbf{Qualitative ablation results for 19 keyframes of human hand puppetry video targets.}}
	\label{fig:paper_alphabet_qualitative}
\end{figure}

\begin{figure}[t!]
	\centering
	\includegraphics[width=1\textwidth]{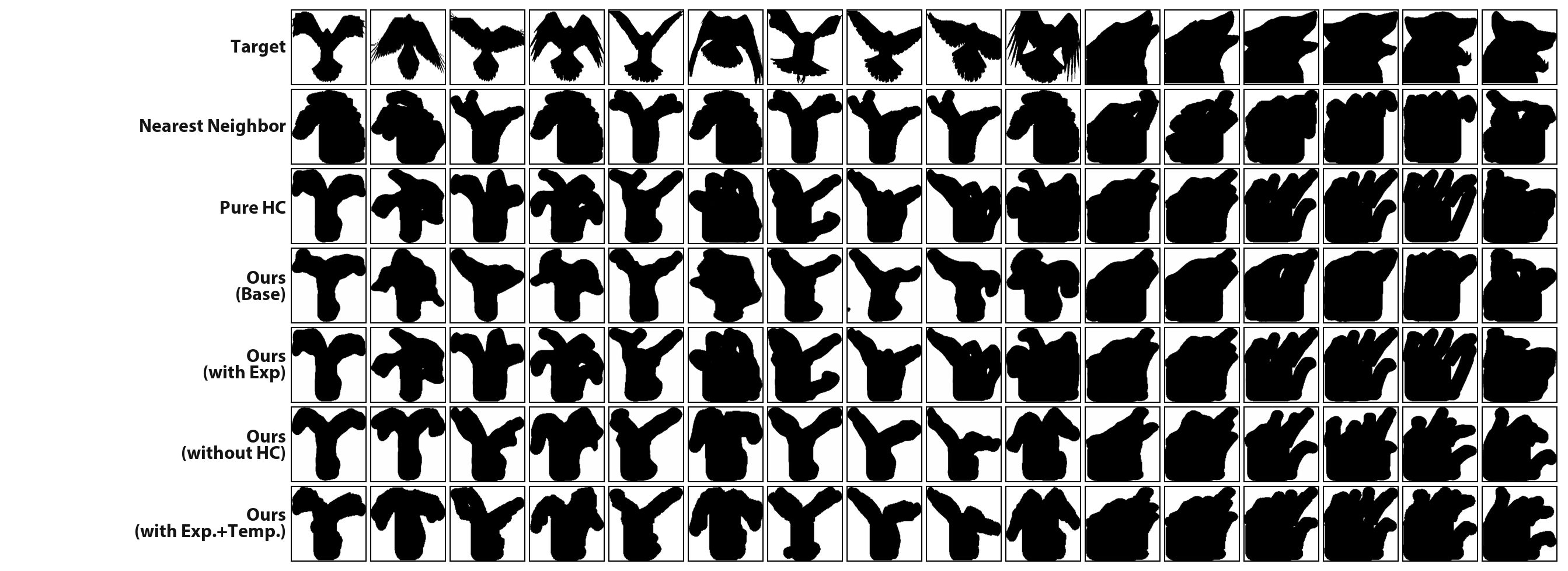} 
	\caption{\textbf{Qualitative ablation results for 16 keyframes of raw animal video targets.}}
	\label{fig:paper_alphabet_qualitative}
\end{figure}
\clearpage
\begin{figure}[t!]
	\centering
	\includegraphics[width=1\textwidth]{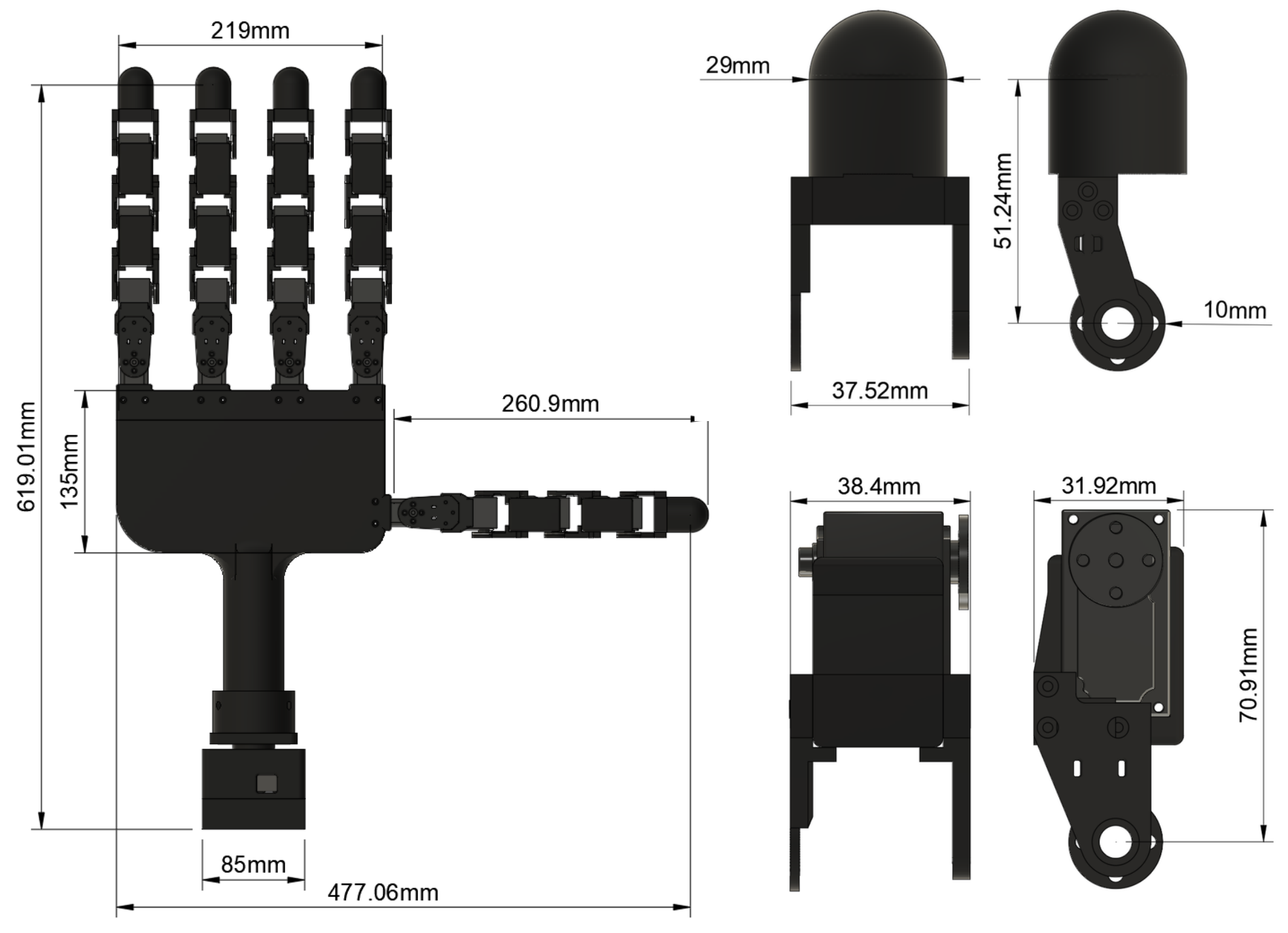} 
	\caption{\textbf{Dimension of the robot skeleton.}}
	\label{fig:dimension}
\end{figure}

\clearpage

\subsection*{Captions for Movies}

\paragraph{Movie 1 | Overview of robotic system.}
The overview of the paper.

\paragraph{Movie S1 | 26 hand gesture shadow imitation}
Imitation of the 26 American Sign Language(ASL) hand gesture shadow.

\paragraph{Movie S2 | Animal shadow puppetry imitation - Deer}
Imitation of the deer shadow puppetry performed by human hand.

\paragraph{Movie S3 | Animal shadow puppetry imitation - Peacock}
Imitation of the peacock shadow puppetry performed by human hand.

\paragraph{Movie S4 | Animal shadow puppetry imitation - Duck}
Imitation of the duck shadow puppetry performed by human hand.

\paragraph{Movie S5 | Animal shadow puppetry imitation - Camel}
Imitation of the camel shadow puppetry performed by human hand.

\paragraph{Movie S6 | Raw animal video imitation - Wolf howling}
Imitation of a raw wolf howling video.

\paragraph{Movie S7 | Raw animal video imitation - Raven flipping wings}
Imitation of a raw raven flipping wings video.

\clearpage 





\end{document}